\pdfoutput=1
\documentclass[11pt]{article}

\usepackage[final]{acl}

\usepackage{times}
\usepackage{latexsym}

\usepackage[T1]{fontenc}
\usepackage[utf8]{inputenc}

\usepackage{microtype}

\usepackage{inconsolata}

\usepackage{graphicx}
\usepackage[utf8]{inputenc} % allow utf-8 input
\usepackage[T1]{fontenc}    % use 8-bit T1 fonts
\usepackage{hyperref}       % hyperlinks
\usepackage{url}            % simple URL typesetting
\usepackage{booktabs}       % professional-quality tables
\usepackage{amsfonts}       % blackboard math symbols
\usepackage{nicefrac}       % compact symbols for 1/2, etc.
\usepackage{microtype}      % microtypography
\usepackage{xcolor}         % colors
\usepackage{enumitem}
\usepackage{amsmath}
\usepackage{times}
\usepackage{booktabs}
\usepackage{latexsym}
\usepackage{graphicx}
\usepackage{multirow}
\usepackage{multicol}
\usepackage{hyperref}
\usepackage{cleveref}
\usepackage{listings}
\usepackage{amsmath}
\usepackage[most]{tcolorbox}
\usepackage{float}
\usepackage{xspace}
\usepackage{lipsum}
\usepackage{subfigure}
\usepackage{enumitem}
\usepackage{amssymb}
\usepackage{pifont}

\title{ProSA: Assessing and Understanding the Prompt Sensitivity of LLMs}

\author{Jingming Zhuo$^{1,2,*}$, Songyang Zhang$^{1,*}$, Xinyu Fang$^{1,3}$, Haodong Duan$^{1}$ \\ {\bf Dahua Lin$^{1,4}$ , Kai Chen$^{1,\dagger}$ } \\
        $^{1}$Shanghai AI Laboratory, $^{2}$Jilin University, $^{3}$Zhejiang University, \\$^{4}$The Chinese University of Hong Kong\\ \texttt{jingmingzhuo@gmail.com, zhangsongyang@pjlab.org.cn} \\
        \small{* equal contribution, $\dagger$ corresponding author}} %\\

\begin{document}
\maketitle
\begin{abstract}
Large language models (LLMs) have demonstrated impressive capabilities across various tasks, but their performance is highly sensitive to the prompts utilized. This variability poses challenges for accurate assessment and user satisfaction. Current research frequently overlooks instance-level prompt variations and their implications on subjective evaluations. To address these shortcomings, we introduce \textbf{ProSA}, a framework designed to evaluate and comprehend prompt sensitivity in LLMs. ProSA incorporates a novel sensitivity metric, \texttt{PromptSensiScore}, and leverages decoding confidence to elucidate underlying mechanisms. Our extensive study, spanning multiple tasks, uncovers that prompt sensitivity fluctuates across datasets and models, with larger models exhibiting enhanced robustness. We observe that few-shot examples can alleviate this sensitivity issue, and subjective evaluations are also susceptible to prompt sensitivities, particularly in complex, reasoning-oriented tasks. Furthermore, our findings indicate that higher model confidence correlates with increased prompt robustness. We believe this work will serve as a helpful tool in studying prompt sensitivity of LLMs. The project is released at: \url{https://github.com/open-compass/ProSA}.

\end{abstract}

\section{Introduction}

In recent years, large language models (LLMs)  have rapidly become the focus of the artificial intelligence field. By training on large-scale corpora, LLMs have shown impressive capabilities in multiple tasks~\citep{zhao2023survey, min2023recent}.
The input for LLMs, known as prompts, is crucial for their ability to complete a wide variety of tasks. 
% By using diverse prompts as input, LLMs can accomplish a range of tasks. 
The prompts for LLMs come in various forms. Even for the same requirement, different individuals' varying expression habits can result in prompts of different styles. Figure 1 illustrates four styles of prompts used by LLMs when completing mathematical tasks.

% The prompt for LLMs, is typically formed by concatenating an \textit{instruction} and an \textit{input}~\citep{alpaca}. The \textit{instructions} specify the task that the model needs to execute and the \textit{inputs} specify the problem's content. 
% If necessary, \textit{complements} such as in-context examples~\citep{brown2020language} can also be incorporated to improve the instruction-following or inspire the model's capabilities. With diverse prompts, LLMs can accomplish various tasks.
The diversity of prompts elicits various responses from LLMs. Recent studies \citep{zhu2023promptbench, pezeshkpour2023large, sclar2023quantifying, zhou2023don} investigate model performance across different prompt templates and demonstrate that LLMs are highly sensitive to the nuances of prompts. Even minor alterations to prompts can lead to substantial declines in model performance. This sensitivity to prompts poses a challenge for researchers aiming to precisely evaluate the models' capabilities. Moreover, users frequently have to iterate on their prompts numerous times to achieve higher quality outputs.

% The diversity in prompts results in a variety of responses from LLMs. Recent studies~\citep{zhu2023promptbench, pezeshkpour2023large} analyse model performance under various instructions and shows that the LLMs are sensitive to prompts.
% With minor prompt changes, the model performance can deteriorate significantly. 
% Such prompt sensitivity makes it challenging for researchers to accurately assess the model's capabilities. Furthermore, users often need to adjust the prompts multiple times to obtain higher-quality responses.
% Such sensitivity presents challenges to robust model training~\citep{zhuo2023robustness} and accurate model assessment~\citep{chang2023survey, liu2023trustworthy}. 
However, current research on prompt sensitivity mainly centers on dataset-level analyses, focusing on performance variations across distinct prompt templates within identical datasets. 
% This limitation constrains the practical relevance of the findings. 
Also, existing works neglect examining how LLMs align with human expectations under different prompts in subjective evaluations, hindering the accuracy of insights reflecting real-world user experiences. Moreover, it's remain changeling in elucidating the reasons behind LLMs' sensitivity to prompts.
% , improving the understanding of the fundamental mechanisms at play. 

% However, existing research on prompt sensitivity primarily focuses on dataset-level prompt sensitivity by evaluating shifts in performance across different instructions within the same dataset, which restricts the practical applicability of the findings. Additionally, they does not investigate how the alignment of LLM responses with human needs varies under different prompts in subjective evaluation, making existing analytical conclusions not accurately reflect the real-world user experience. Furthermore, existing studies do not explain why LLMs exhibit sensitivity to prompts, limiting the understanding of the underlying mechanisms. Addressing these gaps requires conducting a more detailed analysis, conducting analysis on the subjective evaluation, and exploring the reasons behind prompt sensitivity. This comprehensive approach will enhance our understanding of prompt sensitivity and help develop more robust and user-friendly LLMs.

\begin{figure}[!t]
    \centering
    \includegraphics[width=1\columnwidth]{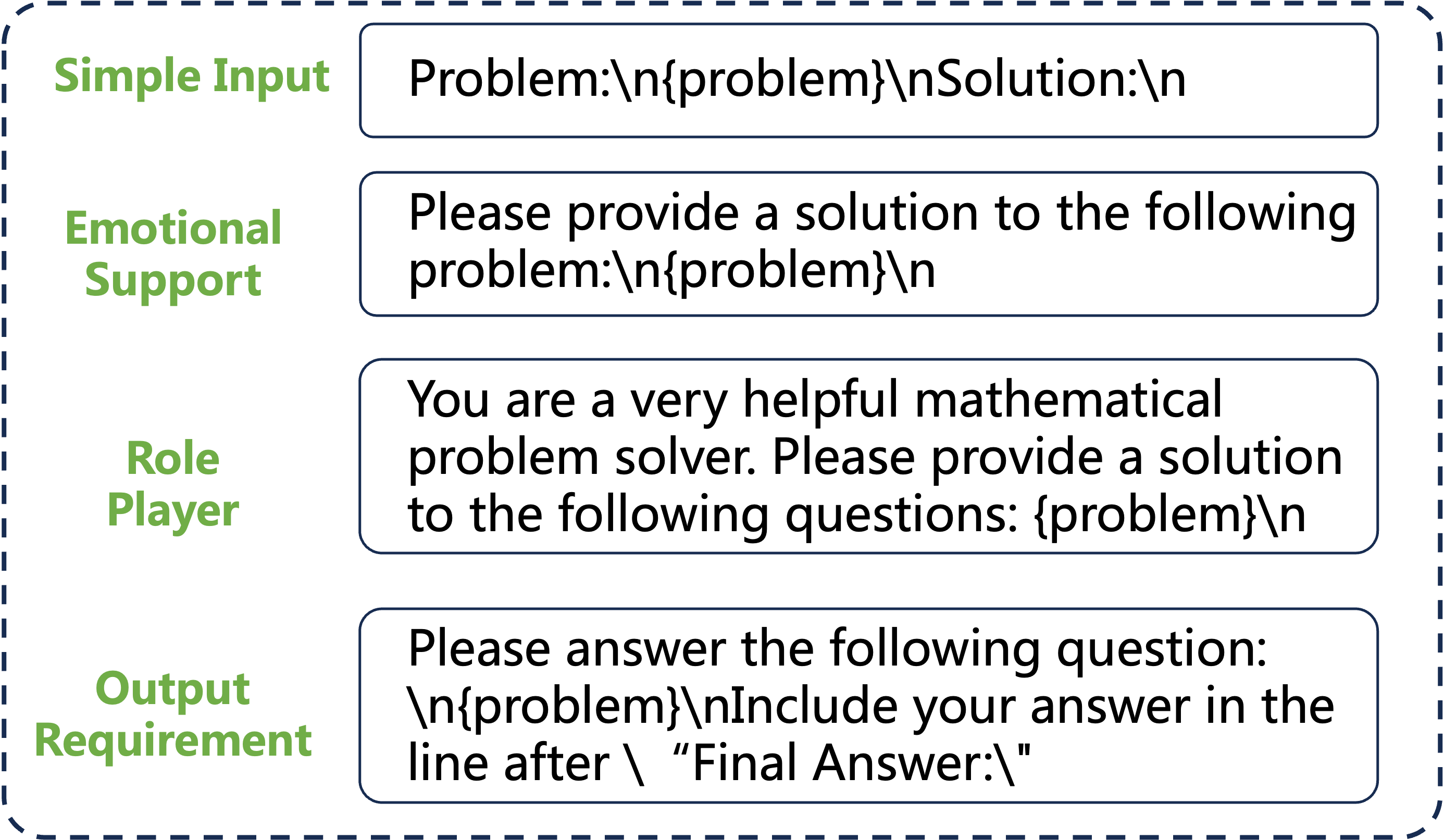}
    \caption{\textbf{A Showcase of the Four Prompt Templates on MATH.} These four prompt templates represent four different styles of constructing prompts, serving as an example of the diversity in human prompt expression.}
    \label{fig:enter-label}
\end{figure}

% They lack a focus on subjective evaluation, failing to investigate how the alignment of LLM responses with human needs varies under different prompts. This oversight can lead to models that do not adequately meet user needs, reducing user satisfaction and trust. Additionally, these studies do not explain why LLMs exhibit sensitivity to prompts, limiting the understanding of the underlying mechanisms and hindering efforts to improve prompt robustness. Furthermore, the analysis is often confined to dataset-level, not considering different application contexts, which restricts the practical applicability of the findings. Addressing these gaps requires emphasizing subjective evaluations, exploring the reasons behind prompt sensitivity, and expanding the scope of analysis to include diverse real-world scenarios. This comprehensive approach will enhance our understanding of prompt sensitivity and help develop more robust and user-friendly LLMs.

% However, many of existing research on prompt sensitivity primarily concentrates on benchmark level analyzes which assess the prompts sensitivity of LLMs, where prompt sensitivity is measured by shifts in the performance of LLMs on diverse benchmarks. Their research also did not involve analyzing the differences in how LLMs' responses align with human needs under different prompts. Furthermore, existing research does not explain why LLMs exhibit sensitivity to prompts.
% This lack leads to a limited understanding of prompt sensitivity and hinders efforts to enhance prompt robustness.

To tackle the aforementioned issues, we introduce the \textbf{ProSA}, focusing on evaluation and understanding the prompt sensitivity of the current LLMs. It can serve as a proxy on monitoring the robustness and stability of the LLMs.
% a comprehensive framework designed for  assessing  and analysing the prompt sensitivity of LLMs.
In this work, we devise the ProSA by focusing on the \textit{instance-level} analysis, develop a novel \textit{sensitivity metric}, and leveraging \textit{decoding confidence} for elucidating the underlying mechanisms. Specifically, we emphasize instance-level measurements and analyses, comprising both objective evaluations against precise references and open-ended subjective evaluations. Our experimental scope spans multiple tasks, ranging from understanding and logical reasoning to coding and general alignment capabilities. 

% Toward this goal, introduce a new metric to measure the prompt sensitivity, which focuses on the inconsistant of the model

To achieve this, we introduce a novel metric designed to quantify prompt sensitivity.  
% Initially, we construct a gallery of semantic prompt variants following the predefined criteria.  
We define our metric, termed \texttt{PromptSensiScore} or PSS, as  the average discrepancy in the LLM's responses when confronted with different semantic variants of the same instruction (prompt). A comprehensive explanation of \texttt{PromptSensiScore}~(PSS) is provided in Sec.~\ref{sec:sensiscore}.

Furthermore, we investigate prompt sensitivity by measuring the instance-level PSS of several popular open-source LLMs through the objective evaluation and subjective evaluation. In the objective evaluation, we assess 8 LLMs across 4 datasets, spanning diverse capabilities, with each model tested on 12 prompt variants. Findings reveal variations in prompt sensitivity among models and datasets, with Llama3-70B-Instruct~\citep{llama3modelcard} demonstrating the highest robustness. 
% Notably, smaller models like Llama3-8B-instruct and InternLM2-7B-chat also exhibit high robustness. 
The study further illustrates that incorporating few-shot examples alleviate prompt sensitivity, particularly evident in the transition from zero- to one-shot scenarios. Larger LLMs especially benefit from increased few-shot instances, exhibiting greater robustness improvements.

Additionally, we analyze five advanced LLMs using two prominent subjective evaluation benchmarks: LC AlpacaEval 2.0~\citep{dubois2024length} and Arena Hard Auto~\citep{arenahard2024}. Our findings illustrate that these models demonstrate high resilience in answering straightforward queries but encounter heightened sensitivity with more complex ones. Categorizing prompts reveals that LLMs are particularly robust when drawing upon established knowledge domains, such as troubleshooting IT issues. Conversely, in coding tasks or those requiring creativity, LLMs prove to be more susceptible to variations in prompts.

% Moreover, we analyze five advanced models on two widely used subjective evaluation benchmarks: 
% LC AlpacaEval 2.0 and Arena Hard. The experiments indicate that these LLMs show high robustness in response quality for simple questions, but exhibit higher sensitivity when it comes to difficult questions. Additionally, the analysis of prompt categories reveals that LLMs exhibit robustness to prompts when addressing prior knowledge, such as IT problems. In tasks involving code development or creative tasks, LLMs are more sensitive to prompts.

We also delve into the underlying reasons for prompt sensitivity. Leveraging the objective evaluation as a stand-in, we utilize the decoding confidence associated with the correct answer to scrutinize model behavior. Findings suggest that prompt sensitivity is essentially a reflection of the model's confidence level: higher confidence in its outputs correlates with increased robustness against prompt semantic variations.
% Furthermore, we explore the causes of prompt sensitivity. We take the objective evaluation as the proxy and employ the decoding confidence of the ground truth option to study the behavior. Experimental results show that prompt sensitivity is an external manifestation of the model's confidence. When the model's confidence in its responses is higher, it also exhibits greater robustness to prompts.

Our contributions can be summarized as follows:
\begin{enumerate}[label={\bf {{$\bullet$}}},leftmargin=*,topsep=0.5ex,itemsep=-0.5ex,partopsep=0.75ex,parsep=0.75ex,partopsep=0pt,wide,labelindent=0pt]

\item We introduce \textbf{ProSA}, a comprehensive framework that places emphasis on instance-level analysis, incorporates a novel sensitivity metric, and utilizes decoding confidence to unravel the underlying mechanisms of prompt sensitivity in LLMs.

\item We propose a novel metric, \texttt{PromptSensiScore} (PSS), which quantifies the average discrepancy in LLM responses when faced with different prompt variants of the same instance. The subsequent analysis on the objective and subjective evaluation provided assistance and guidance in exploring prompt sensitivity.

\item We find that prompt sensitivity is fundamentally an outward appearance the model's decoing confidence: greater confidence in its outputs corresponds to enhanced resilience against changes in prompt semantics.

\end{enumerate}

\section{Instance Level Prompt Sensitivity}
\subsection{Definition}
When instructing LLMs to complete the same task, different individuals often use different expressions, which can lead to the LLMs generating different responses. In this paper, we refer to the specific requirements as an instance. Different expressions of an instance are referred to as different prompts. 

Previous research \citep{zhu2023promptbench, pezeshkpour2023large} compares different prompt templates within datasets (composed of instances) to analyze the prompt sensitivity of LLMs, which is done by examining the shifts in the LLMs' scores across different prompt templates within the dataset. This approach to studying prompt sensitivity has certain limitations.

\begin{figure}[!ht]
    \centering
    \includegraphics[width=\linewidth]{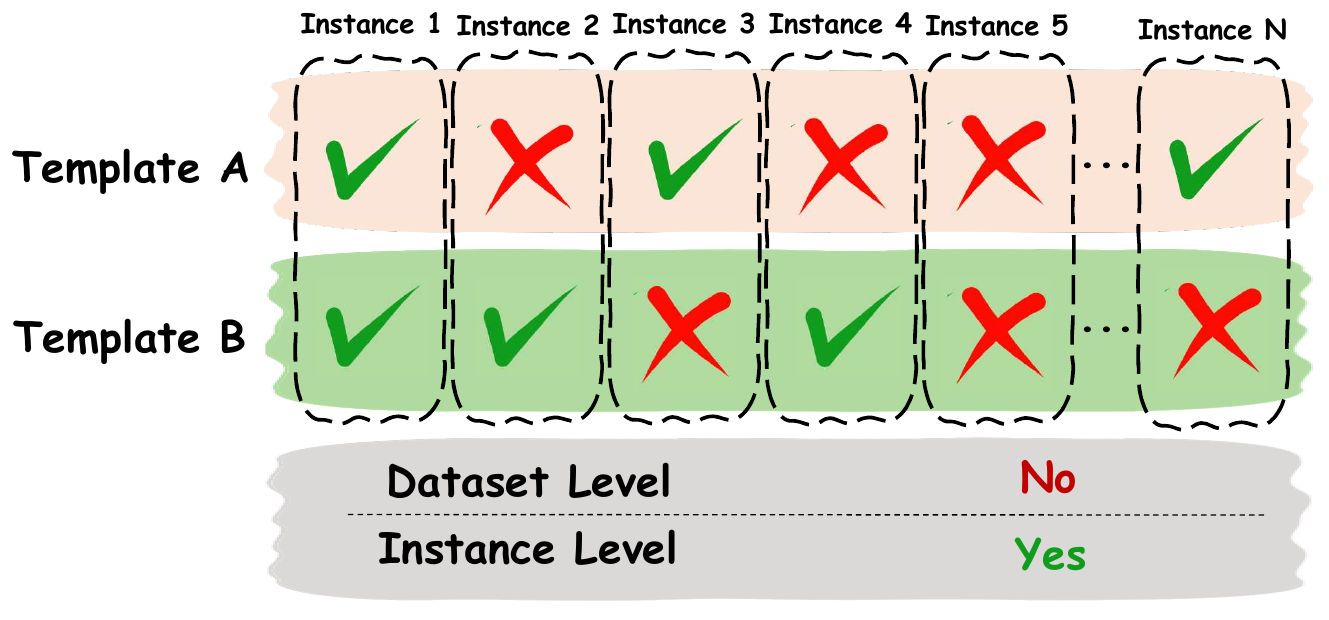}
    \caption{\textbf{A Comparision of Evaluating LLMs' Prompt Sensitivity.} \ding{51} and \ding{55} indicate the accuracy of the LLM's responses. In this example, LLMs appear robust at the dataset level evaluation (calculated from the variance of different templates), but this overlooks the sensitivity of LLMs to different templates within the same instance.}
    \label{fig:sensitivity}
\end{figure}

Despite all instances in a dataset following the same topic, the differences in model performance under the same instances with different prompt templates are often overlooked. Each instance can vary widely in complexity, context, and information type, ranging from straightforward factual questions to those requiring nuanced understanding. This diversity means the model's sensitivity to prompts can differ significantly between instances. As shown in Figure \ref{fig:sensitivity}, LLMs may be robust to templates for some instances while being sensitive to templates for others.

Additionally, relying on aggregate metrics to assess prompt sensitivity can obscure important nuances. While overall trends may be useful, they can mask differences in individual instances where the model's responses drastically changes due to slight prompt modifications.

Therefore, by analyzing instance level prompt sensitivity, we can gain deeper insights into how various factors influence model responses, leading to more effective and reliable use of LLMs.

\subsection{Evaluation Metric}\label{sec:sensiscore}
To analyze prompt sensitivity from the instance level, rather than dataset level perspective, we defined PSS to measure the sensitivity of LLMs to prompts. For each set of all prompt variants under the same instance, we have:

\begin{equation}
    % PromptScore(Inst, p_i, p_j) =
    % \left\{
    % \begin{array}{ll}
    %   1 - Cons(R(p_i), R(p_j)) & p_i \neq p_j,\\ 
    %   0 & otherwise
    % \end{array} \right.
    \mathcal{S} = \frac{\sum_{p_i, p_j \in P} (\left | Y(P_i) - Y(P_j) \right |)}{C(\left | P \right |, 2)}
\end{equation}
Here, $Y(p)$ represents the performance metric under this prompt $p$. For instances with the given ground truth, $Y(p)$ refers to the correctness of LLM' response. For tasks without explicit ground truth, where responses are often given a score representing the quality of the generation, $Y(p)$ refers to the given score within the interval [0, 1]. $\left | Y(P_i) - Y(P_j) \right |$ represents the absolute value difference in performance metrics between prompt \(p_i\) and prompt \(p_j\). $C(\left | P \right |, 2)$ represents the count of prompt pairs in the same instance. The calculation of PSS is as follows:

\begin{equation}
    \text{PSS} = \frac{1}{N} \sum_{i=1}^{N} \mathcal{S}_i
\end{equation}
Here \( N \) is the total number of instances in the dataset and \( \mathcal{S}_i \) is the score for the \( i \)-th instance.

% Due to the different types of tasks and evaluation methods, PSS has different meanings in objective evaluation and subjective evaluation. In the objective evaluation, PSS represents the probability of inconsistency in correctness between any two prompts for the same instance. In the subjective evaluation, PSS indicates the difference in average response quality between two prompts for the same instance.

Due to the different types of tasks and evaluation methods, PSS has different meanings in objective evaluation and subjective evaluation. In objective evaluation, PSS represents the expected inconsistency in the correctness of the model's responses under two different prompts for the same instance. In subjective evaluations, PSS represents the expected difference in response quality (as scored by advanced LLMs). We generally consider that when PSS is less than 0.1, LLMs exhibit high prompt robustness. However, different tasks have varying requirements for LLMs' prompt sensitivity. For instance, tasks like code assistant may require higher robustness to enhance user experience. On Appendix \ref{sec:pss_examples}, we provide examples of model responses under different prompts along with their PSS scores to offer a clearer understanding.

Compared to the statistical analysis of performance shifts, PSS provides a novel perspective a more accurate and intuitive characterization for measuring and analyzing prompt sensitivity.

\section{Prompt Sensitivity on the Objective Evaluation}
\label{sec:objective}
The objective evaluation is a common form for evaluating LLMs. In this setup, the given tasks often have a specific ground truth, such as the correct options in multiple-choice questions or the answers to mathematical problems. The prompt sensitivity on the objective evaluation is analyzed by extracting the outcome of LLM's different responses under various prompts for the same instance.

\begin{figure*}[!t]
    % \centering
    \includegraphics[width=\linewidth]{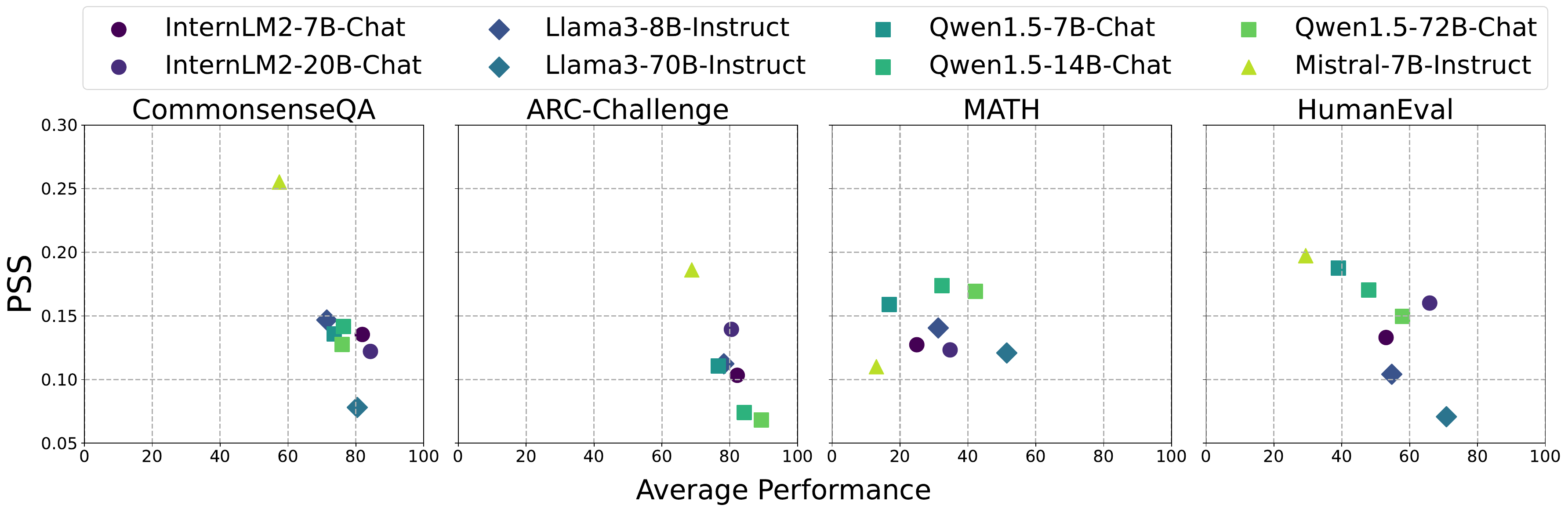}
    \caption{\textbf{Main Results of Prompt Sensitivity.} The scatter represents the average performance score of 12 prompts and the PSS under different datasets.}
    \label{fig:main}
\end{figure*}

\subsection{Experimental Setup}

\noindent \textbf{Dataset Selection}
To comprehensively analyze the prompts sensitivity of LLMs to different forms of task, the datasets for evaluating are as follows:

\begin{enumerate}[label={\bf {{$\bullet$}}},leftmargin=*,topsep=0.5ex,itemsep=-0.5ex,partopsep=0.75ex,parsep=0.75ex,partopsep=0pt,wide,labelindent=0pt]

\item \textbf{CommonsenseQA}~\citep{talmor2018commonsenseqa}: CommonsenseQA is a multiple-choice question dataset with five options per question. It assesses the language model's capability to utilize prior world knowledge.

\item \textbf{ARC-Challenge}~\citep{clark2018think}: ARC-Challenge is a multiple-choice question dataset with four options per question. It consists of elementary science questions, evaluating the reasoning ability of LLMs.

\item \textbf{MATH}~\citep{hendrycks2021measuring}: MATH is a mathematics QA dataset containing different difficulty levels, whose questions are derived from high school math competitions.

\item \textbf{HumanEval}~\citep{chen2021codex}: HumanEval is composed of programming problems constructed by coding experts, designed to evaluate code understanding, simple algorithms, and mathematics.

\end{enumerate}

We adopt the 0-shot setting to evaluate LLMs on the CommonsenseQA, ARC-Challenge and Humaneval. For MATH, we adopt the widely used 4-shot setting.
Additionally, inspired by OpenAI's simple-evals\footnote{https://github.com/openai/simple-evals}, we use Llama3-70B-Instruct to help extract LLM responses for MATH, alleviating bias from incorrect extraction of model responses\footnote{All experiments are conducted with OpenCompass~\citep{2023opencompass}}. The details are shown on Appendix \ref{sec:extract}. 

% Additionally, to minimize bias arising from incorrect extraction of model responses, we use Llama3-70B-Instruct to assist us in extracting the responses of LLMs for MATH.

\noindent \textbf{LLMs Selection}
To comprehensively investigate the sensitivity of LLMs to various prompts, we conducted experiments on a wide range of LLMs with varying sizes, 
including: Llama3 series~\citep{llama3modelcard}, Qwen1.5 series~\citep{qwen}, InternLM2 series~\citep{cai2024internlm2}, and Mistral-7B-Instruct~\citep{jiang2023mistral}. we use greedy decoding in inference to ensure that the results are reproducible.

\noindent \textbf{Prompts Selection}
In real-world scenarios, different users often use different prompt words to indicate the same intent. To align with the richness of human expression, we start with four constructive aspects, including \texttt{Simple Inputs}, \texttt{Role Player}, \texttt{Emotional Support}, and \texttt{Output Requirement}. For each aspect, we have 3 manually constructed prompts with high quality. Figure \ref{fig:enter-label} provides a showcase of four aspects of prompts used for the MATH Dataset. More details about instances and prompts are shown on Appendix \ref{b_1} and \ref{c_1}.

% \begin{figure}
%     \centering
%     \includegraphics[width=1\linewidth]{figure/showcase.png}
%     \caption{\textbf{A showcase of the used prompts from different aspects.} These prompts are used in the MATH benchmark to simulate the diverse expression styles of humans.}
%     \label{fig:enter-label}
% \end{figure}

\subsection{Main Results and Analysis}
We report the main results of the prompt sensitivity on the objective evalution in Figure \ref{fig:main}.

% \subsubsection{Prompt Sensitivity Presents Different Phenomena on Different Datasets}
Due to differences in task types and difficulties, Prompt Sensitivity exhibits varying phenomena across different Datasets. For relatively easier Datasets like CommonsenseQA, ARC-Challenge and HumanEval, the LLMs' average performances and PSS appear to have an approximately linear relationship, indicating that LLMs can achieve high performance while maintaining low prompt sensitivity. In the case of MATH, since it requires extensive reasoning processes to arrive at the correct answers, all LLMs exhibited poor performance along with a certain degree of prompt sensitivity.

% \subsubsection{Prompt sensitivity Lacks Consistency across Datasets}
One LLM may show high sensitive on one task but be robust to prompts on another task. 
For instance, Qwen1.5-14B-Chat is robust to prompts on CommmonsenseQA, but it has the most serious prompt sensitivity on MATH among all LLMs.

We also evaluate two proprietary models on the HumanEval. Due to space constraints, more analysis will be presented on Appendix \ref{sec:proprietary}.

\begin{figure}[!h]
    % \centering
    \includegraphics[width=\columnwidth]{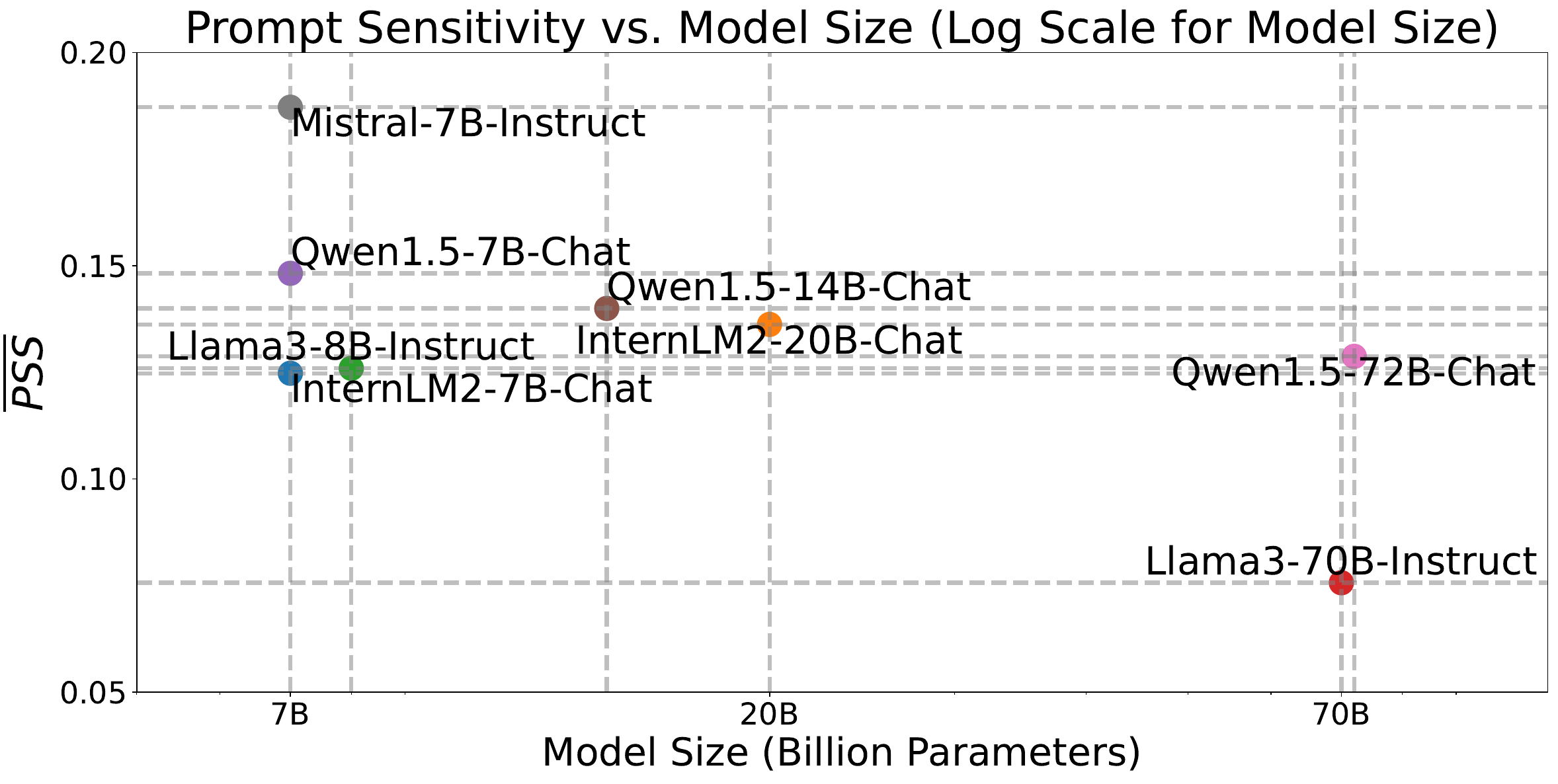}
    \caption{\textbf{Prompt Sensitivity vs. Model Size.} The comparative charts display the relationship between the size of the model's parameters and prompts sensitivity. $\overline{PSS}$ refers to the average PSS of four datasets.}
    \label{fig:size}
\end{figure}

\subsection{Prompt Sensitivity and Model Size}

A commonly raised concern is identifying which LLMs are more sensitive to prompts and whether there is a correlation between a model's prompt sensitivity and its size. For model $l$, we calculate $\overline{PSS}$, the average PSS on the four tested datasets to measure the performance of LLMs in terms of prompt sensitivity.

% \begin{equation}
%     \overline{SensiScore} = \frac{\sum_{d\in \mathcal{D}}SensiScore(d)}{|\mathcal{D}|}
% \end{equation}

% Here, $\mathcal{D}$ denotes the set of tested datasets~(CommonSenseQA, ARC-Challenge, MATH, and HumanEval).

As shown in Figure \ref{fig:size}, LLama3-70B-Instruct demonstrates exceptional robustness to prompts.
After that, the InternLM2-7B-Chat and Llama3-8B-Instruct also showcase notable robustness even though they have a relatively small size, which is counter-intuitive. The three models in the same series, Qwen1.5-7B-Chat, Qwen1.5-20B-Chat, and Qwen1.5-72B-Chat, maintain relatively similar prompt sensitivities while differing greatly in the model sizes. 

\begin{figure*}[!t]
    \centering
    \includegraphics[width=0.93\linewidth]{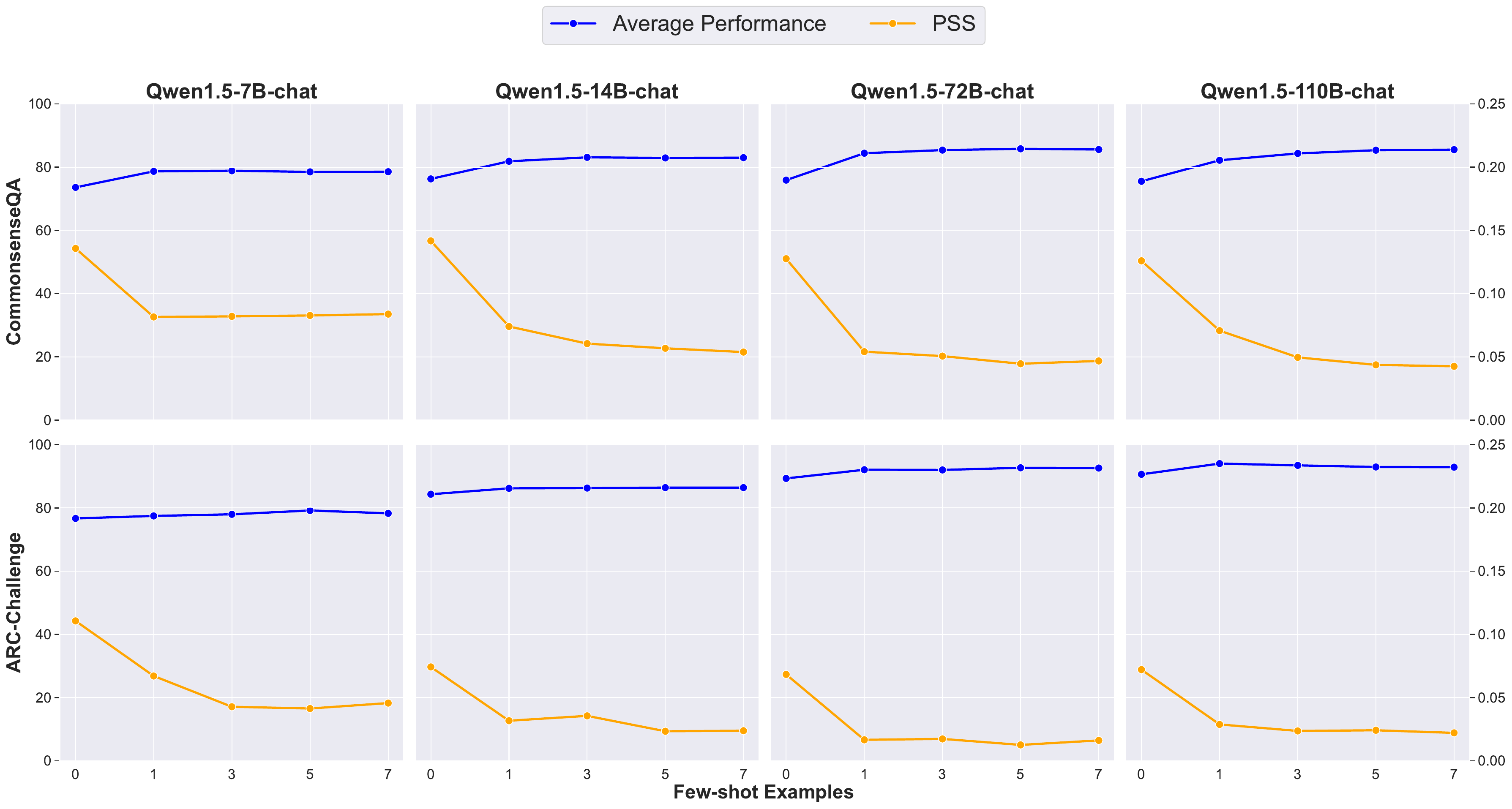}
    \caption{\textbf{Impact of Few-shot on the Performance and Sensitivity.} Conduct experiments on the CommonsenseQA and ARC-Challenge datasets using five few-shot settings and four models from the Qwen series. The blue line represents the changes in the scores of LLMs (using the left scale). The orange line represents the changes in the PSS of LLMs (using the right scale).}
    \label{fig:icl}
\end{figure*}

\subsection{Few-shot Enhances Prompt Robustness}
In the era of LLMs, few-shot plays a critical role in enabling LLMs to follow specific formats and improve their performance~\citep{brown2020language}. To investigate the impact of few-shot on the prompt sensitivity of LLMs, we conduct experiments utilizing the CommonsenseQA and ARC-Challenge.

Specifically, we use the same 12 prompts mentioned earlier to conduct experiments on four models from the Qwen1.5 series, which range in size from 7B to 110B. For each model, we employ greedy decoding. In terms of the number of shot examples, we compare and analyze the 0-shot, 1-shot, 3-shot, 5-shot, and 7-shot methods. In all settings, the first \(k\) shots are the same (if applicable), to mitigate the impact of exaple selection. The few-shot examples used are shown on Appendix \ref{sec:few-shot}.

As shown in \Cref{fig:icl}, the introduction of few-shot learning enhances the robustness of the model's prompts for all models. On the ARC-Challenge, even though the inclusion of few-shot learning did not significantly improve the model's performance, there was a noticeable decrease in the model's sensitivity to prompts. This reduction in sensitivity is most pronounced when transitioning from the 0-shot setting to the 1-shot setting. 
% Additionally, although in-context examples can reduce the model's sensitivity to prompts, the model's preference for prompt presence persists even with an increase in few-shot examples.

% For all LLMs, increasing the number of few-shot examples results in improved robustness for different prompts. 
% The most significant improvement is observed when moving from zero-shot to one-shot. 
% Additionally, the gap between the scores of different prompts narrowed, and the LLMs became more robust to prompts with an increase in in-context examples. However, although in-context examples can reduce the model's sensitivity to prompts, the performance of LLMs still varies across prompts, which means that the model's preference for prompt presence persists even with an increase in few-shot examples.

Besides, as the number of few-shot examples increases, larger LLMs exhibit more robust behavior to prompts. For example, the four models have similar prompt sensitivity under the zero-shot setting on the CommonsenseQA. However, with the increase in few-shot examples, the larger models demonstrate a trend of being more robust to prompts. The same phenomenon is observed on the ARC-Challenge. As the number of few-shot examples increases, Larger models can exhibit better prompt robustness.

Due to limitations in computational resources, we are unable to investigate the effects of using a dozen or more few-shot examples on the sensitivity of small or large models, like whether small models can continue to improve their robustness with an increasing number of few-shot examples, or if this is a capability unique to large models. However, we believe that this direction is worth further exploration.

% \subsection{Differences under Prompt tracks}

\section{Prompt Sensitivity on the Subjective Evaluation}
With the emergence of LLMs' capabilities and their large-scale deployment as assistances or tools serving humans, how to evaluate the quality of LLMs' responses is receiving increasing attention. In real-world scenarios, the vast majority of instances do not have a specific ground truth that is definitively better than others. Thus, some previous work has attempted to evaluate the quality of generated responses using the subjective evaluation. This evaluation is often carried out by either using human raters or a powerful model, such as GPT-4, to score the generated text. This subjective evaluation, compared to the objective evaluation, better reflects the alignment of LLMs with human needs. However, to the best of our knowledge, existing research on prompt sensitivity has not involved an analysis of prompt sensitivity on the subjective evaluation.

\subsection{Experimental Setup}
\subsubsection{Dataset Selection}
Due to the high costs associated with using humans to evaluate LLMs' responses, existing evaluations often use strong judge LLMs to score responses instead of human annotators and it has been demonstrated that the scores from judge LLMs have high consistency with human ratings. In this study, we selected two widely used benchmarks for our experiments:

\begin{enumerate}[label={\bf {{$\bullet$}}},leftmargin=*,topsep=0.5ex,itemsep=-0.5ex,partopsep=0.75ex,parsep=0.75ex,partopsep=0pt,wide,labelindent=0pt]

\item \textbf{LC AlpacaEval 2.0}~\citep{dubois2024length}: LC AlpacaEval 2.0 is a length controlled version of AlpacaEval, which mitigates the judge LLM's preference for longer responses. LC AlpacaEval 2.0 uses GPT-4-1106-preview as the LLM evaluator, scoring the responses of the tested model by comparing them with the reference responses from gpt-4-turbo. A simple Generalized Linear Model (GLM) is then used to correct for the length preference of the LLM evaluator. The tested model's responses will be given a score in the range of [0, 1]. LC AlpacaEval 2.0 consists of a total of 805 instances.

\item \textbf{Arena Hard Auto}~\citep{arenahard2024}: Arena Hard Auto is a benchmark designed to clearly distinguish model capabilities and reflect human preferences. It consists of 250 question categories, with each category containing two questions, forming a total of 500 test questions. Arena Hard Auto also uses a comparative method to score the quality of responses from two models. For responses from Model A and Model B, the LLM evaluator assigns one of five labels: A >> B, A > B, A ~= B, B > A, or B >> A.\footnote{To conduct the sensitivity analysis experiment, we mapped these five labels to 0, 0.25, 0.5, 0.75, and 1.0, respectively, instead of using the win-loss relationship for ELO battles as in the original setup.} To eliminate the positional bias of the LLM evaluator, Arena Hard Auto scores each pair of responses twice, swapping their positions, resulting in a total of 1000 scores.

\end{enumerate}

For both benchmarks, we use the default evaluator, GPT-4-1106-preview, as the LLM evaluator and the default comparison model responses as references (The default versions are GPT-4-1106-preview for LC AlpacaEval 2.0 and GPT-4-0314 for Arena Hard Auto). Appendix \ref{b_2} provides some examples of both benchmarks.

\subsubsection{LLMs Selection}
Given that the two benchmarks score the tested models by comparing their responses with those of the reference models, we selected five models with relatively better performance for our experiments. The selected models are: InternLM2-20B-Chat, Llama3-8B-Instruct, Llama3-70B-Instruct, Qwen1.5-14B-Chat, and Qwen1.5-72B-Chat.

\subsubsection{Prompt Rewriting}
For both benchmarks, we used LLMs to rewrite all the prompts. To achieve a richer variety of question styles, we used two powerful models for the rewriting: GPT-4o and GPT-4-0409. We then manually verified and refined the rewritten questions to ensure their quality. More details about prompt for constructing promtps and cases of rewritten prompts are shown on Appendix \ref{c_se} and \ref{construct}.
% The prompts used for rewriting and the quality verification process for the rewritten questions will be detailed in Appendix.

\begin{table}[!h]
\centering
\resizebox{1 \columnwidth }{!}{%
\begin{tabular}{l|cc|cc}
\hline
\multirow{2}{*}{\textbf{Generator}} & \multicolumn{2}{c|}{\textbf{LC AlpacaEval 2.0}} & \multicolumn{2}{c}{\textbf{Arena Hard Auto}} \\
                           & \textbf{BS}             & \textbf{HS}               & \textbf{BS}               & \textbf{HS}               \\
\hline
GPT-4o                     &  0.94              & 0.89            & 0.94            & 0.92            \\
GPT-4-0409                 &  0.95              & 0.91            & 0.93            & 0.88           \\
\hline
\end{tabular}
}
\caption{\textbf{Verifications for Rewritten Prompts.} Here, \textbf{BS} stands for BERTScore, and \textbf{HS} stands for Human-labeled Similarity.}
\label{tab:verification}
\end{table}

% We conducted two quality verifications on the generated prompts. First, we used the "all-MiniLM-L6-v2" model to calculate the BERTScore~\citep{zhang2019bertscore} (F1 version) between the original and generated prompts. Besides, we recruited five proficient English human annotators to conduct a binary classification task to assess the similarity between the original and generated prompts. For each benchmark, We selected 100 prompt pairs for human annotation. As shown in Table \ref{tab:verification}, the generated texts performed well in both rule-based and human-based quality verifications, demonstrating high semantic similarity between the generated prompts and the original prompts.

We conducted two quality verifications on the rewritten prompts. First, we used the "all-MiniLM-L6-v2" model from Sentence Transformers~\citep{reimers-2019-sentence-bert} to calculate the BERTScore~\citep{zhang2019bertscore} (F1 version) between the original and rewritten prompts.
Besides, we conduct a human verification to assess the similarity between the original and rewritten prompts. We selected 100 prompt pairs for human annotation for each benchmark. For each prompt pair, we had proficient English annotators assess the similarity based on whether the two prompts convey the same semantics, differing only in expression style, to complete a binary classification task. As shown in Table \ref{tab:verification}, the generated texts performed well in both rule-based and human-based quality verifications, demonstrating high semantic similarity between the rewritten prompts and the original prompts.

\begin{table}[!h]
\centering
\resizebox{1 \columnwidth }{!}{%
\begin{tabular}{l|c|c}
\hline
\textbf{Benchmarks} & \textbf{LC AlpacaEval 2.0} & \textbf{Arena Hard Auto} \\
\hline
\textbf{Reference} & 0.167 & 0.275 \\
\hline
InternLM2-20B-Chat & 0.022 & 0.249 \\
% \hline
Llama3-8B-Instruct & 0.013 & 0.266 \\
% \hline
Llama3-70B-Instruct & 0.016 & 0.258 \\
% \hline
Qwen1.5-14B-Chat & 0.022 & 0.249 \\
% \hline
Qwen1.5-72B-Chat & 0.036 & 0.250 \\
\hline
\end{tabular}
}
\caption{\textbf{PSS on LC AlpacaEval 2.0 and Arena Hard Auto.} Reference refers to the average quality difference of responses generated by Llama3-8b-Instruct and Llama3-70b-Instruct. The others represent the PSS of LLMs under the three prompt versions (One original and two generated). Due to the different default comparison models, the PSS of LC AlpacaEval 2.0 and Arena Hard cannot be directly compared.}
\label{tab:subjective}
\end{table}

\subsection{Main Results and Analysis}
We calculated the PSS under three versions (one original and two generated) on two benchmarks, respectively. Since PSS represents the average response quality difference between two prompts of the same instance, we used the average response quality of Llama3-8B-instruct and Llama3-70B-Instruct as references.

\begin{figure*}[!t]
    % \centering
    \includegraphics[width=2.0\columnwidth]{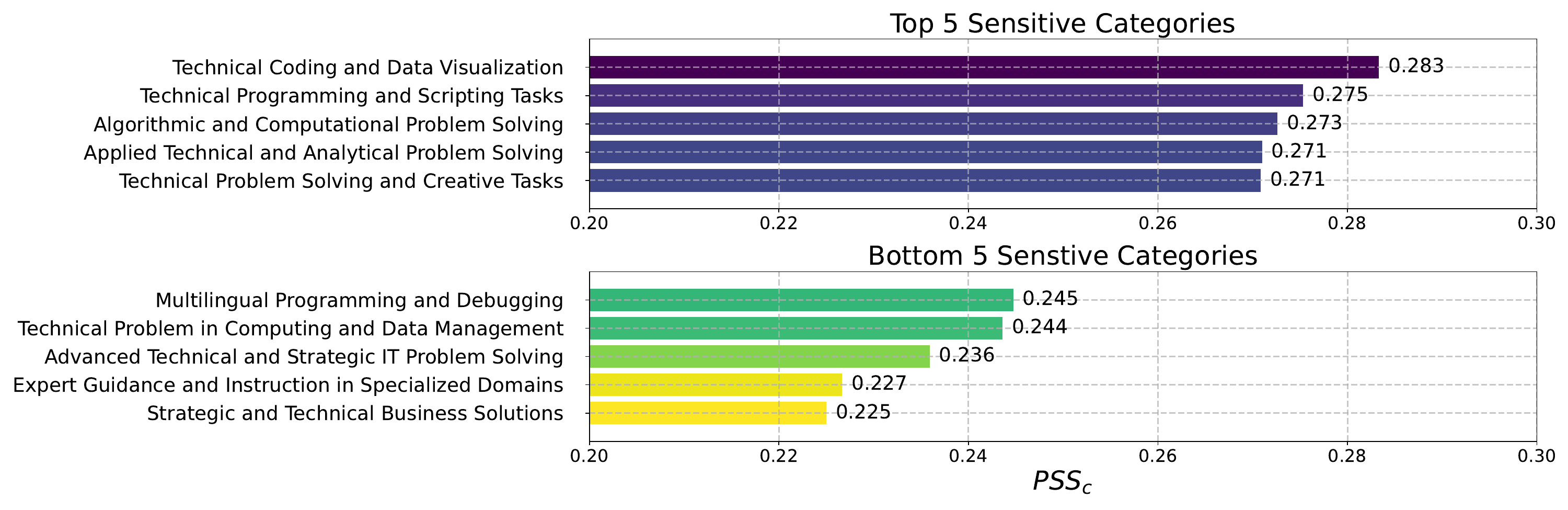}
    \caption{\textbf{Prompt Sensitivity of Different Categories on Arena Hard Auto.} We separately presented the five most sensitive and the five least sensitive categories on Arena Hard Auto. The $PSS_c$ for a particular category refers to the average of the PSS of five LLMs in that category.}
    \label{fig:rank}
\end{figure*}

As shown in Table \ref{tab:subjective}, all LLMs have significantly lower PSS on LC AlpacaEval 2.0 compared to the Reference, demonstrating a certain degree of prompt robustness on this benchmark. However, on Arena Hard Auto, LLMs have demonstrated higher sensitivity. This indicates that this five advanced LLMs have achieved good robustness on relatively simple questions, but they still exhibit prompt sensitivity on more challenging questions.

\subsection{Prompt Sensitivity and Categories}
Users should have a psychological expectation when using LLMs, understanding that these models can be sensitive to prompts. This awareness allows users to modify prompts multiple times to achieve better responses for different tasks. Therefore, measuring the relationship between prompt sensitivity and task categories is valuable. We conduct experiments on Arena Hard Auto to explore the relationship between prompt sensitivity and the category of prompts. 

Due to the excessive complexity of the original 250 categories in Arena Hard Auto and the high randomness of having only 2 instances in each category, we have re-clustered them. We referred to the clustering method used in LMSYS-CHAT-1M~\citep{zheng2023lmsys}. First, we obtained the sentence embeddings for 500 prompts using the "all-MiniLM-L6-v2" model. Then, we applied the k-means algorithm to obtain 20 categories.
We used GPT-4o to name the resulting categories. Subsequently, we calculated the PSS between each of the 20 categories. Figure \ref{fig:rank} shows the five categories with the highest and lowest PSS, respectively.

As shown in Figure \ref{fig:rank}, LLMs exhibit varying degrees of prompt sensitivity across different categories. We can observe that LLMs generally demonstrate better robustness for tasks that heavily test the model's knowledge rather than reasoning, such as business solutions and IT problem. However, when the prompts shift to tasks involving data visualization and scripting tasks, which requires LLMs to generate a large amount of content and involves a certain degree of creativity, LLMs exhibit a higher sensitivity.
\section{Why LLMs are Sensitive to Prompts}
Although the sensitivity of LLMs to prompts significantly impacts user experience and model evaluation, existing research mainly focuses on selecting a suitable prompt from a large set of prompts, rather than involving the interpretation of prompt sensitivity. To address this, we conducted experiments on the CommonsenseQA dataset using Mistral-7B-Instruct, InternLM2-7B-Chat, and InternLM2-20B-Chat. Our experimental results revealed that a model's prompt sensitivity is related to its confidence. For an instance, the more confident the model has, the more robust it is to the prompts.

\subsection{Decoding Confidence}

We use Token Probabilities, a widely employed method for measuring a model's confidence in its generated content~\citep{schuster2022confident, geng2023survey}, to calculate the model's decoding confidence. 
Since CommonsenseQA is a multiple choice question dataset, we calculate the probabilities of only one token predicted by the model for the options. The confidence under an instance is defined as follows:
\begin{equation}
\mathcal{C} = \frac{\sum_{p \in P} P(t_{next} \mid p)}{|P|}
\end{equation}
Here, $P(t_{next} \mid p)$ represents the probability of the token with the max probability predicted by the model under the prompt $p$ in the prompt set $P$. The confidence of LLMs is the average of $\mathcal{C}$ for instances.

\subsection{Experiments and Analysis}

\begin{figure}[!h]
    \centering
    \includegraphics[width=\columnwidth]{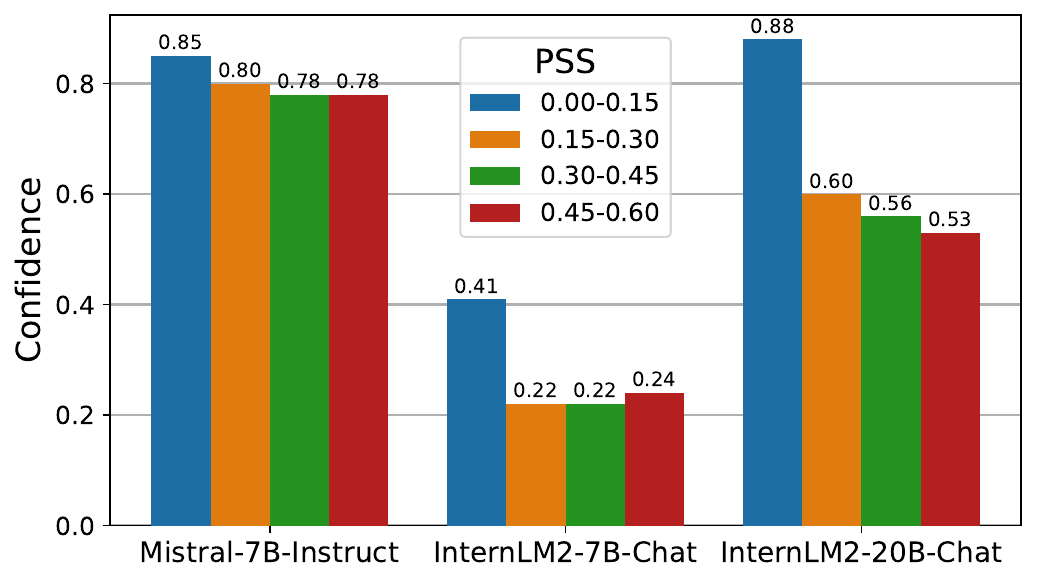}
    \caption{\textbf{The Relationship between Model Confidence and Prompt Sensitivity on CommonsenseQA.} Each bar represents the model's average confidence when its PPS falls within that interval. Note that due to variations in model and vocabulary size, cross-model confidence comparisons are not meaningful.}
    \label{fig:confidence}
\end{figure}

We calculated the decoding confidence of Mistral-7B-Instruct, InternLM2-7B-Chat, and InternLM2-20B-Chat using the same 12 prompt templates as in Section \ref{sec:objective}. The results are shown in Figure \ref{fig:confidence}.

For each LLM, when the model is robust to prompts for a given instance, indicated by a low PPS score, it exhibits the highest decoding confidence. Conversely, when the model is sensitive to prompts for the same instance, its decoding confidence decreases accordingly. This correlation between the model's prompt sensitivity and decoding confidence suggests that prompt sensitivity is an external manifestation of the model's decoding logic.

% As shown in Table \ref{fig:confidence}, when the model selects the same option across twelve prompts, indicating robustness to the question, it also demonstrates high confidence. As the number of different options chosen by the model increases, the model's confidence in its responses gradually decreases. This indicates that the model's prompt sensitivity is due to varying levels of confidence in different questions. When the model has lower confidence in a particular question, it generates different responses under different prompts, exhibiting higher prompt sensitivity.

\section{Related Work}
\subsection{LLMs Evaluation}
Evaluating LLMs is crucial for their efficient use and continuous improvement.
Prior research has systematically unraveled the multifaceted capabilities of LLMs, employing a variety of tasks to assess their performance from different perspectives. These specific tasks include, but are not limited to, reading comprehension~\citep{sakaguchi2019winogrande, mostafazadeh2017lsdsem}, mathematical problem solving~\citep{cobbe2021gsm8k, hendrycks2021measuring}, and code generation~\citep{chen2021codex, austin2021program}. This analytical approach enables a nuanced understanding of how LLMs perform across different facets, shedding light on their efficacy and potential areas for improvement. Recently, with the improvement in the objective capabilities of LLMs, subjective evaluation, which aims to assess the alignment of generated responses with human needs, such as LC AlpacaEval 2.0~\citep{dubois2024length}, has received increasing attention.

\subsection{Prompt Sensitivity}
Previous study~\citep{zhu2023promptbench, pezeshkpour2023large, sclar2023quantifying} showed that LLMs are sensitive to prompts, and that perturbing the prompt can cause a significant variation in the performance of models. \cite{pezeshkpour2023large} demonstrated that LLMs are sensitive to the order of options in multiple choice questions. \cite{mizrahi2023state} demonstrated that model robustness leads to cherry-picking of model performance. However, existing research on prompt sensitivity is insufficient. 
Their research primarily focuses on dataset-level prompt sensitivity analysis. Moreover, existing prompt sensitivity analyses do not address subjective evaluation benchmarks.
Furthermore, the aforementioned research does not address the issue at its root, specifying how to obtain a more robust model.

% \noindent \textbf{Instruction Tuning} Fine-tuning on a large number of instructions aligns the model with human needs and improves the model's performance~\citep{peng2023instruction, zhang2023instruction}.
% Previous work has constructed a large amount of data in terms of quantity and difficulty, and fine-tuned the model to excel in performance~\citep{alpaca, chiang2023vicuna, xu2023wizardlm}. However, these models remain sensitive to prompts, and often only specific prompts can motivate their best performance. Therefore, analyzing and exploring ways to enhance the robustness of the models is crucial.
% \begin{figure}[!h]
%     \centering
%     \includegraphics[width=0.8\columnwidth]{figure/confidence_final.pdf}
%     \caption{\textbf{The Relationship between Model Confidence and Prompt Sensitivity on CommonsenseQA.} Each bar represents the model's average confidence when its PPS falls within that interval. Note that due to variations in model and vocabulary size, cross-model confidence comparisons are not meaningful.}
%     \label{fig:confidence}
% \end{figure}

\section{Conclusion}
In summary, we propose an instance-level prompt sensitivity metric, PSS, and conduct a comprehensive analysis on both objective and subjective evaluation. Additionally, we explore the relationship between prompt sensitivity and model confidence. We believe our work can provide guidance for further sensitivity analysis and building robust LLMs.
\section{Limitations}
In this work, we evaluate the sensitivity of LLMS to prompts on both objective and subjective evaluation, but we also recognize the shortcomings of our work. Due to the limitation of computational resources, we can't explore the phenomenon of prompt sensitivity as the examples of fow-shot increase. In addition, due to the high price of OpenAI API service, we only conducted experiments on three prompt variants on LC AlpacaEval 2.0 and Arena Hard Auto.
\section{Ethical Considerations}
We utilize publicly available datasets and LLMs for our analytical experiments. While we acknowledge that our findings could potentially be used to create cherry-picked evaluation reports, we believe our work will contribute to enhancing the robustness of LLMs to various prompts. Additionally, we employ GPT-4o to refine our writing and assist us to create figures.
\section{Acknowledge}
This paper is supported by the National Key R\&D Program of China (No. 2022ZD0161600).

\bibliography{custom}

\appendix

% \section{Example Appendix}
% \label{sec:appendix}
% \clearpage
\definecolor{lightgray}{gray}{0.95}
\definecolor{codegreen}{rgb}{0,0.6,0}
\definecolor{codegray}{rgb}{0.5,0.5,0.5}
\definecolor{codepurple}{rgb}{0.58,0,0.82}
\definecolor{maroon}{cmyk}{0,0.1,0.01,0.01}
\definecolor{blue}{cmyk}{0.95,0.0,0.2,0.2}
\definecolor{yellow}{cmyk}{0.01,0.0,0.2,0.01}
\definecolor{lightblue}{cmyk}{0.1,0.0,0.02,0.02}
\definecolor{orangeyellow}{cmyk}{0,0.6,1,0}
\lstdefinestyle{prompt}{
    basicstyle=\ttfamily\fontsize{7pt}{8pt}\selectfont,
    frame=none,
    breaklines=true,
    backgroundcolor=\color{lightgray},
    breakatwhitespace=true,
    breakindent=0pt,
    escapeinside={(*@}{@*)},
    numbers=none,
    numbersep=5pt,
    xleftmargin=5pt,
}

\lstdefinestyle{code}{
    % language=Basic,
    basicstyle=\ttfamily\fontsize{7pt}{8pt}\selectfont,
    keywordstyle=\color{magenta},
    commentstyle=\color{codegreen},
    stringstyle=\color{codepurple},
    showstringspaces=false，
    breaklines=true,
    backgroundcolor=\color{lightgray},
    breakatwhitespace=true,
    breakindent=0pt,
    numbers=none,
    numbersep=5pt,
    xleftmargin=5pt,
}
\lstdefinestyle{myprompt}{
    % basicstyle=\ttfamily\fontsize{7pt}{8pt}\selectfont,
    frame=none,
    breaklines=true,
    breakatwhitespace=true,
    breakindent=0pt,
    escapeinside={(*@}{@*)},
    numbers=none,
    % numbersep=5pt,
    % xleftmargin=5pt,
}
\tcbset{
  aibox/.style={
    top=10pt,
    colback=white,
    colframe=black,
    colbacktitle=black,
    fontlower=\ttfamily,
    enhanced,
    center,
    % breakable,
    attach boxed title to top left={yshift=-0.1in,xshift=0.15in},
    boxed title style={boxrule=0pt,colframe=white,},
  }
}
\newtcolorbox{AIbox}[2][]{aibox, title=#2,#1}

\section{Examples of Model Responses and PSS}
\label{sec:pss_examples}
Examples of model responses and PSS are shown in Table \ref{tab:pss_examples}, where the "Responses" column indicates the correctness of the model's responses under 12 different prompt templates. A list of 12 elements represents the responses of the LLM to the 12 prompt templates for one specific instance. A value of 1 represents a correct response, while 0 indicates an incorrect response.
\begin{table}[!h]
\centering
\resizebox{1 \columnwidth }{!}{%
\begin{tabular}{c|c|c}
\toprule
\textbf{Examples} & \textbf{Responses} & \textbf{PSS} \\ \midrule
1 & [1, 1, 1, 1, 1, 1, 1, 1, 1, 1, 1, 1] & 0 \\ \hline
2 & [1, 1, 1, 0, 1, 1, 1, 1, 1, 1, 1, 1] & 0.17 \\ \hline
3 & [1, 1, 0, 0, 1, 1, 0, 1, 1, 1, 1, 1] & 0.41 \\ \bottomrule
\end{tabular}
}
\caption{\textbf{Examples of Model Responses and PSS.} This table provides three examples of what the PSS values are for given responses.}
\label{tab:pss_examples}
\end{table}

\section{Details about Extracting Answers from MATH}
\label{sec:extract}
Due to the open-domain response patterns and the equivalence of different mathematical formats (for example, 0.5 and \textbackslash frac\{1\}\{2\} are equivalent), it is not easy to accurately validate the correctness of responses in MATH.

Instead of relying on the common method of extracting answers from the "\textbackslash box\{\}", we follow OpenAI's simple-evals, utilizing advanced LLMs (in this case, Llama3-70B-Instruct) for post-processing to extract answers. We use the prompts in Figure \ref{fig:extract} to Prompt LLMs to complete the answer extraction task. This approach of using advanced LLMs for post-processing allows us to more accurately assess the LLMs' responses, thereby making our experimental analysis more reliable.

\section{Examples of the Datasets}
\subsection{Objective Evaluation}
\label{b_1}
We selected four representative instances in ARC-Challenge and MATH as examples for the presentation of evaluation data and prompts.
The examples are shown in the \Cref{fig:arc_exp_1,fig:arc_exp_2,fig:arc_exp_3,fig:arc_exp_4} and \Cref{fig:math_exp_1,fig:math_exp_2,fig:math_exp_3,fig:math_exp_4}.

\subsection{Subjective Evaluation}
\label{b_2}
We selected four representative instances in LC AlpacaEval 2.0 and Arena Hard Auto as examples for the presentation of evaluation data and prompts.
The examples are shown in the \Cref{fig:alpacaeval_exp_1,fig:alpacaeval_exp_2,fig:alpacaeval_exp_3} and \Cref{fig:arenahard_exp_1,fig:arenahard_exp_2,fig:arenahard_exp_3}.

\section{Prompt Templates of Each Datasets}
\subsection{Objective Evaluation}
\label{c_1}
For each objective evaluation dataset, we generated a total of 12 prompts to interrogate the same instance in different ways to examine the model's prompt sensitivity. The respective prompts for each dataset are shown in \Cref{fig:arc_prompt_template,fig:csqa_prompt_template,fig:math_prompt_template,fig:humaneval_prompt_template}.
\subsection{Subjective Evaluation}
\label{c_se}
Since the subjective datasets were all requested to be generated by the model based on certain requirements, we rewrote the original prompt using GPT-4 to obtain two new prompts to be evaluated separately. Here (\Cref{fig:alpacaeval_prompt_template,fig:arenahard_prompt_template}) we give two examples for each dataset to show the difference between the prompts before and after rewriting.

\section{Results and Analysis about the Proprietary Models}
\label{sec:proprietary}
Here, we present the results of the two most powerful proprietary LLMs, Claude-3.5-sonnet and GPT-4o, on the HumanEval benchmark with 12 prompts, and compare them with several advanced open-source LLMs, as shown on Table \ref{tab:propietary}.

\begin{table}[!h]
\centering
\resizebox{1 \columnwidth }{!}{%
\begin{tabular}{l|c|c}
\toprule
LLMs & PSS & Avg Acc. \\ \midrule
Claude-3.5-sonnet & 0.14 & 76.37 \\ \hline
GPT-4o & 0.15 & 79.78 \\ \hline
Llama3-8B-Instruct & 0.10 & 54.73 \\ \hline
Llama3-70B-Instruct & 0.07 & 70.83 \\ \hline
Qwen1.5-72B-Instruct & 0.15 & 57.88 \\ \bottomrule
\end{tabular}
}
\caption{\textbf{Results about Several Models on Humaneval.}}
\label{tab:propietary}
\end{table}

The experimental results indicate that even though Claude-3.5-sonnet and GPT-4o achieve optimal performance on average, they are still not considered robust to prompts compared to other models with lower scores. This suggests that prompt sensitivity remains an important research topic. We believe that the ProSA framework can facilitate further research on prompt sensitivity and inspire the design of more robust LLMs.

We conducted a case study on GPT-4o and Llama3-72B-Instruct. We found that for instances where GPT-4o or Llama-3-70B-Instruct had a higher proportion of correct answers, they were able to respond correctly across all 12 prompt templates. However, when presented with instances that appeared to be more challenging for them, the situation changed. Even when different prompt templates were used, Llama-3-70B-Instruct still tended to answer incorrectly on instances it struggled with. In contrast, GPT-4o often exhibited a different behavior: it could provide correct answers under certain prompts even when it had previously answered incorrectly under others. This differing behavior is a key reason for the observed disparity in their prompt sensitivity. This phenomenon has motivate us to reflect on the selection between "unstable geniuses" and "stable mediocrities."

\section{Few-shot Examples of the Objective Datasets}
\label{sec:few-shot}
The few-shot examples of ARC-Challenge and CommonseQA are given in \Cref{fig:arc_7shot,fig:csqa_7shot}, respectively.

\section{Prompts to GPT-4 when Rewriting Prompts in the Subjective Evaluation}
\label{construct}
The prompts used to GPT-4 when rewriting prompts in the subjective evaluation are presented in the \Cref{fig:meta_prompt}.

% #########

\begin{figure*}[!h] 
\begin{AIbox}{Prompt for Extracting Answers form MATH}
{
Look at the following two expressions (answers to a math problem) and judge whether they are equivalent. Only perform trivial simplifications.\\
\noindent\par
... (Few shot examples)\\
\noindent\par
- - -\\
YOUR TASK\\
Respond with only "Result: [[Correct]]" or "Result: [[Incorrect]]" (without quotes). Do not include a rationale.\\
\makebox[2em][l]{\ \ }Expression 1: \{obj\_gold\}\\
\makebox[2em][l]{\ \ }Expression 2: \{prediction\}\\
}
\end{AIbox} 
\caption{\textbf{Prompt for Extracting Answers form MATH.}}
\label{fig:extract}
\vspace{-5mm}
\end{figure*}

\begin{figure*}[!ht] 
% \vspace{1mm}
\begin{AIbox}{ARC-Challenge Example I}
{
{\color{red}\texttt{Prompt:}} One year, the oak trees in a park began producing more acorns than usual. The next year, the population of chipmunks in the park also increased. Which best explains why there were more chipmunks the next year? \\
A. Shady areas increased. \\
B. Food sources increased. \\
C. Oxygen levels increased. \\
D. Available water increased. \\
Answer: \\
{\color{orangeyellow}\texttt{Ground Truth Answer:}} B. Food sources increased. \\
{\color{blue}\texttt{InternLM2-7B-Chat Answer:}} B. Food sources increased.
}
\end{AIbox} 
\caption{\textbf{Example I of the ARC-Challenge.}}
\label{fig:arc_exp_1}
\vspace{-5mm}
\end{figure*}

\begin{figure*}[!ht] 
\begin{AIbox}{ARC-Challenge Example II}
{
{\color{red}\texttt{Prompt:}} Could you provide a response to the following question: Scientific models are very common. For which purpose below would a physical model be least helpful? A. simulating scientific phenomena B. simplifying a complex idea C. allowing visualization D. displaying data \\
{\color{orangeyellow}\texttt{Ground Truth Answer:}} D. displaying data \\
{\color{blue}\texttt{InternLM2-7B-Chat Answer:}} A physical model would be least helpful for the purpose of \"displaying data.\" While physical models can be used to display data, they are not the most efficient or effective method for this purpose. Displaying data typically refers to presenting information in a clear and organized manner, often through the use of graphs, charts, or tables. Physical models are better suited for simulating scientific phenomena, simplifying complex ideas, and allowing visualization, as they can provide a tangible representation of a concept or system that can be manipulated and observed.
}
\end{AIbox} 
\caption{ \bf Example II of the ARC-Challenge.}
\label{fig:arc_exp_2}
\end{figure*}

\begin{figure*}[!ht] 
% \vspace{1mm}
\begin{AIbox}{ARC-Challenge Example III}
{
{\color{red}\texttt{Prompt:}} As an exceptionally resourceful AI assistant, I'm at your service. Address the questions below:\\
The Humboldt River Basin is one of the most important sources of several nonrenewable resources. Which of these nonrenewable resources is produced from the Humboldt River Basin?\\
A. coal from mountain deposits\\
B. oil and gas from underground reservoirs\\
C. electric energy from hydroelectric generators\\
D. minerals from mining gold, silver, and copper \\
{\color{orangeyellow}\texttt{Ground Truth Answer:}} D. minerals from mining gold, silver, and copper \\
{\color{blue}\texttt{InternLM2-7B-Chat Answer:}} The Humboldt River Basin is primarily known for its hydroelectric potential, with several dams and power plants utilizing its water for electricity generation. Therefore, the correct answer is C. electric energy from hydroelectric generators. While the basin does contain some mineral deposits, such as gold, silver, and copper, these are not considered nonrenewable resources in the same way that fossil fuels are. Additionally, the Humboldt River Basin does not have significant coal deposits or oil and gas reservoirs.
}
\end{AIbox} 
\caption{ \bf Example III of the ARC-Challenge.}
\label{fig:arc_exp_3}
\end{figure*}

\begin{figure*}[!ht] 
\vspace{-5mm}
\begin{AIbox}{ARC-Challenge Example IV}
{
{\color{red}\texttt{Prompt:}} Please answer the following question:\\
Which describes the composition of carbohydrates?\\
A. lipids bonding to form phospholipids\\
B. monomers bonding to form polymers\\
C. amino acids bonding to form polypeptides\\
D. saccharides bonding to form polysaccharides\\
Answer the question by replying A, B, C or D.\\
{\color{orangeyellow}\texttt{Ground Truth Answer:}} D. saccharides bonding to form polysaccharides \\
{\color{blue}\texttt{InternLM2-7B-Chat Answer:}} B. monomers bonding to form polymers
}
\end{AIbox} 
\caption{ \bf Example IV of the ARC-Challenge.}
\label{fig:arc_exp_4}
\vspace{-5mm}
\end{figure*}

\begin{figure*}[!ht] 
\vspace{-5mm}
\begin{AIbox}{MATH Example I}
{
{\color{red}\texttt{Prompt:}} Problem: 
Determine the remainder of 194 (mod 11).\\
Solution: \\
{\color{orangeyellow}\texttt{Ground Truth Answer:}} 7 \\
{\color{blue}\texttt{Llama3-8B-Instruct Answer:}} To find the remainder of 194 when divided by 11, we can use the fact that $194 = 11(17) + 7$. Therefore, the remainder is $\boxed{7}$.\\
Final Answer: The final answer is 7.
}
\end{AIbox} 
\caption{ \bf Example I of the MATH.}
\label{fig:math_exp_1}
\end{figure*}

\begin{figure*}[!ht] 
\vspace{-5mm}
\begin{AIbox}{MATH Example II}
{
{\color{red}\texttt{Prompt:}} Please provide a solution to the following problem:\\
If $x=720$ and $ax$ is a perfect cube where $a$ is a positive integer, what is the smallest possible value of $\sqrt[3]{ax}$?\\
{\color{orangeyellow}\texttt{Ground Truth Answer:}} 60 \\
{\color{blue}\texttt{Llama3-8B-Instruct Answer:}} Since $x=720=6^3$, we know that $ax$ is a perfect cube if $a$ is a factor of $720$. The smallest possible value of $a$ is $1$, which gives $ax=720$. Therefore, the smallest possible value of $\sqrt[3]{ax}$ is $\sqrt[3]{720}=\boxed{{6}}$.\\
Final Answer: The final answer is 6.
}
\end{AIbox} 
\caption{ \bf Example II of the MATH.}
\label{fig:math_exp_2}
\end{figure*}

\begin{figure*}[!ht] 
\vspace{-5mm}
\begin{AIbox}{MATH Example III}
{
{\color{red}\texttt{Prompt:}} You are a very helpful mathematical problem solver. Please provide a solution to the following questions: In a certain isosceles right triangle, the altitude to the hypotenuse has length $4\sqrt{2}$.  What is the area of the triangle?\\
{\color{orangeyellow}\texttt{Ground Truth Answer:}} 32 \\
{\color{blue}\texttt{Llama3-8B-Instruct Answer:}} Let $x$ be the length of each leg of the isosceles right triangle.  Then the hypotenuse has length $\sqrt{x^2+x^2}=\sqrt{2x^2}=x\sqrt{2}$.  The altitude to the hypotenuse has length $4\sqrt{2}$, so we can set up the equation $$x\sqrt{2}=4\sqrt{2}\Rightarrow x=4.$$The area of the triangle is therefore $$\frac{1}{2}\cdot4\cdot4=\boxed{8}.$$\\
Final Answer: The final answer is 8.
}
\end{AIbox} 
\caption{ \bf Example III of the MATH.}
\label{fig:math_exp_3}
\vspace{-5mm}
\end{figure*}

\begin{figure*}[!ht] 
\vspace{-5mm}
\begin{AIbox}{MATH Example IV}
{
{\color{red}\texttt{Prompt:}} Please help me to address the following question:\\
A $60^\circ$ rotation around the origin in the counter-clockwise direction is applied to $3 \sqrt{3} - 5i.$  What is the resulting complex number?\\
Include your answer in the line after "Final Answer:" \\
{\color{orangeyellow}\texttt{Ground Truth Answer:}} $4 \sqrt{3} + 2i$ \\
{\color{blue}\texttt{Llama3-8B-Instruct Answer:}} A $60^\circ$ rotation around the origin in the counter-clockwise direction is equivalent to multiplying the complex number by $e^{i\pi/3} = \cos(\pi/3) + i\sin(\pi/3) = \frac{1}{2} + \frac{\sqrt{3}}{2}i.$ Therefore, the resulting complex number is $$\left(3\sqrt{3}-5i\right)\left(\frac{1}{2}+\frac{\sqrt{3}}{2}i\right)=\frac{1}{2}(3\sqrt{3}-5i)+\frac{\sqrt{3}}{2}(3\sqrt{3}-5i)i=\boxed{-\frac{5}{2}+3\sqrt{3}i}.$$\\
Final Answer: The final answer is $-\frac{5}{2}+3\sqrt{3}i$.
}
\end{AIbox} 
\caption{ \bf Example IV of the MATH.}
\label{fig:math_exp_4}
\vspace{-5mm}
\end{figure*}

\begin{figure*}[!ht] 
\vspace{-5mm}
\begin{AIbox}{LC AlpacaEval 2.0 Example I}
{
{\color{red}\texttt{Prompt:}} I'm a broke college student living in Silicon Valley. Tell me a joke mocking Silicon Valley and how broke I am simultaneously. \\
{\color{blue}\texttt{InternLM2-20B-Chat Answer:}} Here's a light-hearted joke that combines elements of Silicon Valley and your financial situation:\\
Why did the broke college student in Silicon Valley refuse to invest in the latest tech startup?\\
Because he couldn't afford the "minimum viable purchase"!\\
This joke playfully acknowledges the high cost of living in Silicon Valley and the common perception of startups requiring significant investments. It also highlights the financial struggles of a broke college student who can't afford to participate in the tech industry's latest trends. Remember, it's essential to maintain a respectful and considerate approach when making light of sensitive topics like financial hardship.
}
\end{AIbox} 
\caption{ \bf Example I of the LC AlpacaEval 2.0.}
\label{fig:alpacaeval_exp_1}
\vspace{-5mm}
\end{figure*}

\begin{figure*}[!ht] 
\vspace{-5mm}
\begin{AIbox}{LC AlpacaEval 2.0 Example II}
{
{\color{red}\texttt{Prompt:}} Compose an email to apologize to a customer for a delayed order and provide assurance that the problem has been fixed.\\
{\color{blue}\texttt{InternLM2-20B-Chat Answer:}} Subject: Apology for Delayed Order and Confirmation of Resolved Issue\\
Dear [Customer's Name],\\
I hope this email finds you well. I am writing to express my sincerest apologies for the delay you have experienced with your recent order, [Order Number]. We understand the inconvenience this has caused you and we are truly sorry for any frustration it may have caused.\\
At [Company Name], we pride ourselves on delivering high-quality products and exceptional customer service. Unfortunately, we fell short in meeting your expectations with this particular order, and for that, we are deeply sorry.\\
The delay was due to an unforeseen issue in our production process, which has since been identified and resolved. We have taken immediate steps to ensure that this issue will not affect any future orders, and we have implemented additional quality control measures to prevent such incidents from occurring again.\\
To make up for the inconvenience, we would like to offer you a [insert compensation offer, e.g., a discount on your next purchase, free shipping, or an extended warranty]. Please find the details of this offer attached to this email.\\
We value your business and your satisfaction is of utmost importance to us. We understand that your time is valuable, and we appreciate your patience and understanding during this period. We assure you that we are committed to providing you with the best possible service moving forward.\\
If you have any further concerns or questions, please do not hesitate to reach out to our customer care team at [Customer Service Email] or [Customer Service Phone Number]. We are always here to help.\\
Once again, we apologize for the delay and any inconvenience it has caused. We appreciate your continued support and loyalty to [Company Name]. We look forward to the opportunity to serve you better in the future.\\
Warm regards,\\
\quad [Your Name]\\
\quad [Your Title]\\
\quad [Company Name]\\
\quad [Company Email]\\
\quad [Company Phone Number]
}
\end{AIbox} 
\caption{ \bf Example II of the LC AlpacaEval 2.0.}
\label{fig:alpacaeval_exp_2}
\vspace{-10mm}
\end{figure*}

\begin{figure*}[!ht] 
\vspace{-5mm}
\begin{AIbox}{LC AlpacaEval 2.0 Example III}
{
{\color{red}\texttt{Prompt:}} Hello, my sister and her friends have invited me to join them in a game of kickball. Could you outline the rules of the game for me so I can avoid being outplayed?\\
{\color{blue}\texttt{InternLM2-20B-Chat Answer:}} Of course! Kickball is a fun and active game that combines elements of soccer and baseball. Here's a brief outline of the rules to help you prepare for the game:\\
**Equipment:**\\
- A round, inflated ball (similar to a soccer ball)\\
- A kickball field with a pitching mound, bases (usually four), and a home plate\\
- Teams: Typically, two teams of 8-12 players each\\
**Objective:**\\
The objective of kickball is to score runs by kicking the ball and running around the bases. The team with the most runs at the end of the game wins.\\
**Gameplay:**\\
1. **Teams:** The game is played between two teams, with each team taking turns being the "kicking" team and the "fielding" team.\\
2. **Pitching:** The game starts with a kickoff, where the pitcher from the fielding team rolls the ball to the kicker from the kicking team. The kicker then kicks the ball as hard as they can, aiming to hit it as far as possible.\\
3. **Kicking:** The kicker must kick the ball with their hands behind their back and cannot run until the ball has been kicked. Once the ball is kicked, the kicker can run to first base. If the ball is caught by a fielder before it hits the ground, the kicker is out.\\
4. **Running the Bases:** The kicker and their teammates can run around the bases in a counterclockwise direction (similar to baseball). They can advance to the next base when the ball is kicked and caught by a fielder, or when the ball is kicked and hits the ground.\\
5. **Fielding:** The fielding team's goal is to get the kicker and their teammates out by catching the ball before it hits the ground or by tagging a runner with the ball while they are between bases.\\
6. **Outs:** A runner is out if they are tagged with the ball while they are between bases, if the ball is caught before it hits the ground, or if the ball is hit out of bounds.\\
7. **Scoring:** A run is scored when a runner successfully reaches home plate. The game ends when all players on the kicking team have had a chance to kick the ball, or when a predetermined number of innings have been played (usually 6 or 7).\\
**Additional Rules:**\\
- If the ball is kicked out of bounds, the runner must return to the base they were on before the kick.\\
- If the ball is kicked and hits a fielder, the runner can advance to the next base if they were on a base when the ball was kicked.\\
- If the ball is kicked and hits a base, the runner can advance to the next base if they were on a base when the ball was kicked.\\
Remember, the rules may vary slightly depending on the specific version of kickball you're playing. It's always a good idea to confirm the rules with your sister and her friends before the game starts. Have fun and enjoy the game!
}
\end{AIbox} 
\caption{ \bf Example III of the LC AlpacaEval 2.0.}
\label{fig:alpacaeval_exp_3}
\vspace{-5mm}
\end{figure*}

\begin{figure*}[!ht] 
\vspace{-5mm}
\begin{AIbox}{Arena Hard Auto Example I}
{
{\color{red}\texttt{Prompt:}} Write a song about catfish in the style of Bob Dylan.\\
{\color{blue}\texttt{Qwen1.5-72B-Chat Answer:}} \\
(Verse 1)\\
Well, I've been down by the river, where the catfish play,\\
In the muddy waters, where the current sways.\\
They slip through the shadows, with their whiskers so fine,\\
Living in the depths, where the secrets intertwine.\\
(Chorus)\\
Oh, catfish, catfish, swimming in the stream,\\
You're a mystery in the heart of a dream.\\
With your scales and your fins, you're a troubadour of the deep,\\
Catfish, catfish, your story's too hard to keep.\\
(Verse 2)\\
Underneath the moonlight, in the heart of the night,\\
The catfish dances, in the river's light.\\
They've seen the world changing, through the ages gone by,\\
Still, they hold on to the secrets that the water hides.\\(Chorus)\\
Catfish, catfish, in the current's embrace,\\You're a symbol of the rhythm of the river's grace.\\With your quiet strength, you navigate the flow,\\Catfish, catfish, your wisdom we should know.\\(Bridge)\\In the delta's embrace, where the blues were born,\\The catfish whispers tales that have been forlorn.\\Of love and loss, and the secrets of the soul,\\In the echo of the current, where the story's told.\\(Verse 3)\\They say the catfish can climb the highest falls,\\Escaping the chains that bind us all.\\A symbol of resilience, a creature of the night,\\Slipping through the shadows, out of sight.\\(Chorus)\\Oh, catfish, catfish, with your soul so free,\\You teach us to navigate life's troubled sea.\\In the depths and the darkness, you find your way,\\Catfish, catfish, lead us through the fray.\\(Outro)\\So here's to the catfish, in their watery home,\\May we learn from their journey, as we roam.\\In the river's melody, we hear their song,\\Catfish, catfish, forever strong.\\
}
\end{AIbox} 
\caption{ \bf Example I of the Arena Hard Auto.}
\label{fig:arenahard_exp_1}
\vspace{-5mm}
\end{figure*}

\begin{figure*}[!ht] 
\vspace{-5mm}
\begin{AIbox}{Arena Hard Auto Example II}
{
{\color{red}\texttt{Prompt:}} Compose a melody using ABC notation in the style of a folk tune..\\
{\color{blue}\texttt{Qwen1.5-72B-Chat Answer:}} X:1\\T:Untitled Folk Tune\\M:4/4\\L:1/8\\K:G\\|:G2A2|B2A2|G2E2|D4|E2F2|G2F2|E2C2|B,4|\\A2B2|c2B2|A2F2|E4|D2E2|F2E2|D2B,2|C4:|\\
}
\end{AIbox} 
\caption{ \bf Example II of the Arena Hard Auto.}
\label{fig:arenahard_exp_2}
\vspace{-5mm}
\end{figure*}

\begin{figure*}[!ht] 
\vspace{-5mm}
\begin{AIbox}{Arena Hard Auto Example III}
{
{\color{red}\texttt{Prompt:}} What steps should I take to protect my home wifi router? \\
{\color{blue}\texttt{Qwen1.5-72B-Chat Answer:}}\\
Protecting your home Wi-Fi router is essential to ensure the security of your network and devices connected to it. Here are some steps you should take:\\
1. **Change the default login credentials**: Most routers come with default usernames and passwords, which are often easily guessable. Change them to a strong, unique combination of letters, numbers, and symbols.\\2. **Update the firmware**: Regularly check for firmware updates from the manufacturer's website. These updates often contain security patches that fix vulnerabilities.\\3. **Use WPA3 or WPA2 encryption**: Ensure your router is using the latest Wi-Fi security protocol, WPA3 if possible, or WPA2 as a minimum. Avoid using WEP, as it is less secure.\\4. **Set a strong Wi-Fi password**: Use a strong password for your Wi-Fi network, just like you would for the router login. Avoid using personal information and make it complex.\\5. **Enable Network Name (SSID) hiding**: Hiding your Wi-Fi network's name can make it less visible to potential attackers, but note that it won't provide complete security.\\6. **Enable MAC address filtering**: This allows you to specify which devices are allowed to connect to your network based on their unique hardware (MAC) addresses.\\7. **Disable remote management**: If you don't need to manage your router from outside your home network, disable this feature to prevent unauthorized access.\\8. **Use a strong password for all connected devices**: Ensure all devices connected to your Wi-Fi have strong passwords as well, especially smart home devices that may have weaker security.\\9. **Regularly update devices**: Keep all connected devices, including computers, smartphones, and tablets, updated with the latest security patches and software updates.\\10. **Consider a firewall**: Some routers have built-in firewalls; make sure they are enabled. You can also use a software firewall on your devices.\\11. **Monitor network activity**: Regularly check your router's logs to see if there are any unusual connections or activities.\\12. **Use a Virtual Private Network (VPN)**: If you need to access the internet from outside your home network, consider using a VPN to encrypt your data.\\By following these steps, you can significantly improve the security of your home Wi-Fi router and the devices connected to it.\\
}
\end{AIbox} 
\caption{ \bf Example III of the Arena Hard Auto.}
\label{fig:arenahard_exp_3}
\vspace{-5mm}
\end{figure*}

% \subsection{The template for different prompt in four different subjective dataset}

\begin{figure*}[!ht] 
\vspace{-5mm}
\begin{AIbox}{ARC-Challenge Prompt Templates}
{
{\color{blue}\bf Prompt 1:}
\begin{lstlisting}[style=code]
{question}\nA. {A}\nB. {B}\nC. {C}\nD. {D}\nAnswer:
\end{lstlisting}
% \tcline
{\color{blue}\bf Prompt 2:}
\begin{lstlisting}[style=code]
Question:\n{question}\nA. {A}\nB. {B}\nC. {C}\nD. {D}\nAnswer:
\end{lstlisting}
{\color{blue}\bf Prompt 3:}
\begin{lstlisting}[style=code]
Question:\n{question} A. {A} B. {B} C. {C} D. {D}\nAnswer:
\end{lstlisting}
{\color{blue}\bf Prompt 4:}
\begin{lstlisting}[style=code]
Could you provide a response to the following question: {question} A. {A} B. {B} C. {C} D. {D}
\end{lstlisting}
{\color{blue}\bf Prompt 5:}
\begin{lstlisting}[style=code]
Please answer the following question:\n{question}\nA. {A}\nB. {B}\nC. {C}\nD. {D}
\end{lstlisting}
{\color{blue}\bf Prompt 6:}
\begin{lstlisting}[style=code]
Please address the following question:\n{question}\nA. {A}\nB. {B}\nC. {C}\nD. {D}\nAnswer:
\end{lstlisting}
{\color{blue}\bf Prompt 7:}
\begin{lstlisting}[style=code]
You are a very helpful AI assistant. Please answer the following questions: {question} A. {A} 
B. {B} C. {C} D. {D}
\end{lstlisting}
{\color{blue}\bf Prompt 8:}
\begin{lstlisting}[style=code]
As an exceptionally resourceful AI assistant, I'm at your service. Address the questions below:\n
{question}\nA. {A}\nB. {B}\nC. {C}\nD. {D}
\end{lstlisting}
{\color{blue}\bf Prompt 9:}
\begin{lstlisting}[style=code]
As a helpful Artificial Intelligence Assistant, please answer the following questions\n{question}
A. {A}\nB. {B}\nC. {C}\nD. {D}
\end{lstlisting}
{\color{blue}\bf Prompt 10:}
\begin{lstlisting}[style=code]
Could you provide a response to the following question: {question} A. {A} B. {B} C. {C} 
D. {D}\nAnswer the question by replying A, B, C or D.
\end{lstlisting}
{\color{blue}\bf Prompt 11:}
\begin{lstlisting}[style=code]
Please answer the following question:\n{question}\nA. {A}\nB. {B}\nC. {C}\nD. {D}\n
Answer the question by replying A, B, C or D.
\end{lstlisting}
{\color{blue}\bf Prompt 12:}
\begin{lstlisting}[style=code]
Please address the following question:\n{question}\nA. {A}\nB. {B}\nC. {C}\nD. {D}\n
Answer this question by replying A, B, C or D.
\end{lstlisting}
}
\end{AIbox} 
\caption{ \bf Prompt Templates for the ARC-Challenge.}
\label{fig:arc_prompt_template}
\vspace{-5mm}
\end{figure*}

\begin{figure*}[!ht] 
\vspace{-5mm}
\begin{AIbox}{CommonsenseQA Prompt Templates}
{
{\color{blue}\bf Prompt 1:}
\begin{lstlisting}[style=code]
{question}\nA. {A}\nB. {B}\nC. {C}\nD. {D}\nE. {E}\nAnswer:
\end{lstlisting}
% \tcline
{\color{blue}\bf Prompt 2:}
\begin{lstlisting}[style=code]
Question:\n{question}\nA. {A}\nB. {B}\nC. {C}\nD. {D}\nE. {E}\nAnswer:
\end{lstlisting}
{\color{blue}\bf Prompt 3:}
\begin{lstlisting}[style=code]
Question:\n{question} A. {A} B. {B} C. {C} D. {D} E. {E}\nAnswer:
\end{lstlisting}
{\color{blue}\bf Prompt 4:}
\begin{lstlisting}[style=code]
Could you provide a response to the following question: {question} A. {A} B. {B} C. {C} D. {D} E. {E}
\end{lstlisting}
{\color{blue}\bf Prompt 5:}
\begin{lstlisting}[style=code]
Please answer the following question:\n{question}\nA. {A}\nB. {B}\nC. {C}\nD. {D}\nE. {E}
\end{lstlisting}
{\color{blue}\bf Prompt 6:}
\begin{lstlisting}[style=code]
Please address the following question:\n{question}\nA. {A}\nB. {B}\nC. {C}\nD. {D}\nE. {E}\n
Answer:
\end{lstlisting}
{\color{blue}\bf Prompt 7:}
\begin{lstlisting}[style=code]
You are a very helpful AI assistant. Please answer the following questions: {question} A. {A} 
B. {B} C. {C} D. {D} E. {E}
\end{lstlisting}
{\color{blue}\bf Prompt 8:}
\begin{lstlisting}[style=code]
As an exceptionally resourceful AI assistant, I'm at your service. Address the questions below:\n
{question}\nA. {A}\nB. {B}\nC. {C}\nD. {D}\nE. {E}
\end{lstlisting}
{\color{blue}\bf Prompt 9:}
\begin{lstlisting}[style=code]
As a helpful Artificial Intelligence Assistant, please answer the following questions\n{question}
A. {A}\nB. {B}\nC. {C}\nD. {D}\nE. {E}
\end{lstlisting}
{\color{blue}\bf Prompt 10:}
\begin{lstlisting}[style=code]
Could you provide a response to the following question: {question} A. {A} B. {B} C. {C} 
D. {D} E. {E}\nAnswer the question by replying A, B, C, D or E.
\end{lstlisting}
{\color{blue}\bf Prompt 11:}
\begin{lstlisting}[style=code]
Please answer the following question:\n{question}\nA. {A}\nB. {B}\nC. {C}\nD. {D}\nE. {E}\n
Answer the question by replying A, B, C, D or E.
\end{lstlisting}
{\color{blue}\bf Prompt 12:}
\begin{lstlisting}[style=code]
Please address the following question:\n{question}\nA. {A}\nB. {B}\nC. {C}\nD. {D}\nE. {E}\n
Answer this question by replying A, B, C, D or E.
\end{lstlisting}
}
\end{AIbox} 
\caption{ \bf Prompt Templates for the CommonsenseQA.}
\label{fig:csqa_prompt_template}
\vspace{-5mm}
\end{figure*}

\begin{figure*}[!ht] 
\vspace{-5mm}
\begin{AIbox}{MATH}
{
{\color{blue}\bf Prompt 1:}
\begin{lstlisting}[style=code]
Problem:\n{problem}\nSolution:\n
\end{lstlisting}
% \tcline
{\color{blue}\bf Prompt 2:}
\begin{lstlisting}[style=code]
{problem}\nSolution:
\end{lstlisting}
{\color{blue}\bf Prompt 3:}
\begin{lstlisting}[style=code]
Problem: {problem}\nSolution:
\end{lstlisting}
{\color{blue}\bf Prompt 4:}
\begin{lstlisting}[style=code]
Could you provide a solution to the following 
question: {problem}\n
\end{lstlisting}
{\color{blue}\bf Prompt 5:}
\begin{lstlisting}[style=code]
Please provide a solution to the following problem:\n{problem}\n
\end{lstlisting}
{\color{blue}\bf Prompt 6:}
\begin{lstlisting}[style=code]
Please address the following problem:\n{problem}
Answer:
\end{lstlisting}
{\color{blue}\bf Prompt 7:}
\begin{lstlisting}[style=code]
You are a very helpful mathematical problem solver. Please provide a solution to the following 
questions: {problem}\n
\end{lstlisting}
{\color{blue}\bf Prompt 8:}
\begin{lstlisting}[style=code]
As an AI expert in math, could you help me to answer the problem below:\n{problem}\nSolution:\n
\end{lstlisting}
{\color{blue}\bf Prompt 9:}
\begin{lstlisting}[style=code]
As a helpful Artificial Intelligence Assistant, please answer the following question.\n{problem}\n
\end{lstlisting}
{\color{blue}\bf Prompt 10:}
\begin{lstlisting}[style=code]
Solve the following problem: {problem}\nPut your answer on its own line after \"Final Answer:\"
\end{lstlisting}
{\color{blue}\bf Prompt 11:}
\begin{lstlisting}[style=code]
Please answer the following question:\n{problem}\nInclude your answer in the line after \"
Final Answer:\"
\end{lstlisting}
{\color{blue}\bf Prompt 12:}
\begin{lstlisting}[style=code]
Please help me to address the following question:\n{problem}\nInclude your answer in the line after 
\"Final Answer:\"
\end{lstlisting}
}
\end{AIbox} 
\caption{ \bf Prompt Templates for the MATH.}
\label{fig:math_prompt_template}
\vspace{-5mm}
\end{figure*}

\begin{figure*}[!ht] 
\vspace{-5mm}
\begin{AIbox}{HumanEval}
{
{\color{blue}\bf Prompt 1:}
\begin{lstlisting}[style=code]
Create a Python script for this problem:\n{prompt}\nResponse:\n
\end{lstlisting}
% \tcline
{\color{blue}\bf Prompt 2:}
\begin{lstlisting}[style=code]
Provide a Python script that solves the following problem:\n{prompt}\n
\end{lstlisting}
{\color{blue}\bf Prompt 3:}
\begin{lstlisting}[style=code]
Complete the following Python code:\n{prompt}
\end{lstlisting}
{\color{blue}\bf Prompt 4:}
\begin{lstlisting}[style=code]
Please provide a self-contained Python script that solves the following problem in a markdown code 
block:\n```\n{prompt}\n```
\end{lstlisting}
{\color{blue}\bf Prompt 5:}
\begin{lstlisting}[style=code]
Could you provide a response to complete the following Python code:\n{prompt}\nResponse:
\end{lstlisting}
{\color{blue}\bf Prompt 6:}
\begin{lstlisting}[style=code]
Please help me to create a Python script for this problem:\n{prompt}\nResponse:\n
\end{lstlisting}
{\color{blue}\bf Prompt 7:}
\begin{lstlisting}[style=code]
You are a very helpful AI assistant. Could you provide a response to complete the following 
Python code:\n{prompt}\nYour response:
\end{lstlisting}
{\color{blue}\bf Prompt 8:}
\begin{lstlisting}[style=code]
As an outstanding AI assistant, please provide a self-contained Python script that solves the 
following problem in a markdown code block:\n```\n{prompt}\n```\n
\end{lstlisting}
{\color{blue}\bf Prompt 9:}
\begin{lstlisting}[style=code]
As an AI expert in coding. Please help me to create a Python script for this problem:\n{prompt}\n
Your response should only contain the code for the function.
\end{lstlisting}
{\color{blue}\bf Prompt 10:}
\begin{lstlisting}[style=code]
Could you provide a response to complete the following Python code:\n{prompt}\nYou need to put the 
script in the following format:\n```python\n# Your response\n```
\end{lstlisting}
{\color{blue}\bf Prompt 11:}
\begin{lstlisting}[style=code]
Please provide a self-contained Python script that solves the following problem in a markdown code 
block:\n```\n{prompt}\n```\nYou have to follow the following format:```python\n# Your script\n```
\end{lstlisting}
{\color{blue}\bf Prompt 12:}
\begin{lstlisting}[style=code]
Please help me to create a Python script for this problem:\n{prompt}\nYour response should only 
contain the code for the function.
\end{lstlisting}
}
\end{AIbox} 
\caption{ \bf Prompt Templates for the HumanEval.}
\label{fig:humaneval_prompt_template}
\vspace{-5mm}
\end{figure*}

\begin{figure*}[!ht] 
\vspace{-5mm}
\begin{AIbox}{LC AlpacaEval 2.0 Rewritten Prompt Examples}
{
{\bf Example 1:} \\
{\color{blue}\bf Original Prompt:} \\
Create a table with the planets of the solar system and their dimensions \\
{\color{blue}\bf Rewrite Prompts:} \\
1. What are the dimensions of each planet in our solar system, and can you organize them in a table for me? \\
2. Construct a chart listing the planets in the solar system along with their sizes. \\

{\bf Example 2:} \\
{\color{blue}\bf Original Prompt:} \\
I've recently started playing the turn-based strategy game Into the Breach. I have read online that the game is considered to have `perfect information'. What is meant by `perfect information' in this context? \\
{\color{blue}\bf Rewrite Prompts:} \\
1. I recently began playing the turn-based strategy game Into the Breach. I read online that the game is described as having `perfect information'. What does `perfect information' mean in this context? \\
2. I began playing the turn-based strategy game Into the Breach recently and came across the term `perfect information' used in discussions about the game. Can you explain what `perfect information' means in this context? \\
}
\end{AIbox} 
\caption{ \bf Rewritten Prompts Examples for the LC AlpacaEval 2.0.}
\label{fig:alpacaeval_prompt_template}
\vspace{-5mm}
\end{figure*}

\begin{figure*}[!ht] 
\vspace{-5mm}
\begin{AIbox}{Arena Hard Auto Rewritten Prompt Examples}
{
{\bf Example 1:} \\
{\color{blue}\bf Original Prompt:} \\
Explain the book the Alignment problem by Brian Christian. Provide a synopsis of themes and analysis. Recommend a bibliography of related reading. \\
{\color{blue}\bf Rewrite Prompts:} \\
1. Discuss the book 'The Alignment Problem' by Brian Christian. Offer a synopsis of its themes and analysis. Suggest a bibliography of related reading.\\
2. Provide an overview of the book 'The Alignment Problem' authored by Brian Christian, including a synopsis of its themes and analysis. Additionally, suggest a bibliography of related readings. \\

{\bf Example 2:} \\
{\color{blue}\bf Original Prompt:} \\
Query an excel table using MySQL to select dram excel table tree species by diameter class, count the number of representation of the diameter class and some volume of the total \\
{\color{blue}\bf Rewrite Prompts:} \\
1. Query an Excel table using MySQL to select tree species by diameter class, count the number of representations of each diameter class, and calculate the total volume. \\
2. How can I use MySQL to query an Excel table, in order to select tree species from a 'dram' table by diameter class, count the occurrences of each diameter class, and sum the total volume? \\
}
\end{AIbox} 
\caption{ \bf Rewritten Prompt Examples for the Arena Hard Auto.}
\label{fig:arenahard_prompt_template}
\vspace{-5mm}
\end{figure*}

\begin{figure*}[!ht] 
\vspace{-5mm}
\begin{AIbox}{Few-shot Examples for the ARC Challenge}
{
{\color{blue}\bf Example 1:}
\begin{lstlisting}[style=code]
An astronomer observes that a planet rotates faster after a meteorite impact. Which is the most 
likely effect of this increase in rotation?
A. Planetary density will decrease.
B. Planetary years will become longer.
C. Planetary days will become shorter.
D. Planetary gravity will become stronger.
Answer: C. Planetary days will become shorter. 
\end{lstlisting}
{\color{blue}\bf Example 2:}
\begin{lstlisting}[style=code]
A group of engineers wanted to know how different building designs would respond during an 
earthquake. They made several models of buildings and tested each for its ability to withstand 
earthquake conditions. Which will most likely result from testing different building designs?
A. buildings will be built faster
B. buildings will be made safer
C. building designs will look nicer
D. building materials will be cheaper
Answer: B. buildings will be made safer 
\end{lstlisting}
{\color{blue}\bf Example 3:}
\begin{lstlisting}[style=code]
The end result in the process of photosynthesis is the production of sugar and oxygen. Which step 
signals the beginning of photosynthesis?
A. Chemical energy is absorbed through the roots.
B. Light energy is converted to chemical energy.
C. Chlorophyll in the leaf captures light energy.
D. Sunlight is converted into chlorophyll.
Answer: C. Chlorophyll in the leaf captures light energy. 
\end{lstlisting}
{\color{blue}\bf Example 4:}
\begin{lstlisting}[style=code]
A physicist wants to determine the speed a car must reach to jump over a ramp. The physicist 
conducts three trials. In trials two and three, the speed of the car is increased by 20 miles 
per hour. What is the physicist investigating when he changes the speed? 
A. the control 
B. the hypothesis statement 
C. the dependent (responding) variable 
D. the independent (manipulated) variable 
Answer: D. the independent (manipulated) variable
\end{lstlisting}
{\color{blue}\bf Example 5:}
\begin{lstlisting}[style=code]
An astronaut drops a 1.0 kg object and a 5.0 kg object on the Moon. Both objects fall a total 
distance of 2.0 m vertically. Which of the following best describes the objects after they 
have fallen a distance of 1.0 m?
A. They have each lost kinetic energy.
B. They have each gained the same amount of potential energy.
C. They have each lost the same amount of potential energy.
D. They have each gained one-half of their maximum kinetic energy.
Answer: D. They have each gained one-half of their maximum kinetic energy.
\end{lstlisting}
{\color{blue}\bf Example 6:}
\begin{lstlisting}[style=code]
Devil facial tumor disease (DFTD) is a disease that is decimating the population of Tasmanian devils. 
The disease passes from one animal to another through bites and is caused by parasites. The 
parasites cause cancerous tumors that spread throughout an infected animal's body and kill it. 
What is the best description of DFTD?
A. a non-infectious, cell-cycle disease
B. an infectious, cell-cycle disease
C. a non-infectious, chronic disease
D. an infectious, chronic disease
Answer: B. an infectious, cell-cycle disease
\end{lstlisting}
{\color{blue}\bf Example 7:}
\begin{lstlisting}[style=code]
A type of small mammal from the mountain regions of the western United States makes its 
home out of piles of rock. During summer months, the mammal places grasses and seeds in protected 
places in the rock piles. Which of the following is the most likely reason for this behavior?
A. to repare for migration before winter
B. to provide warmth during the cold winter months
C. to store food that will be eaten over the winter months
D. to protect the grasses and seeds from decay before winter
Answer: C. to store food that will be eaten over the winter months
\end{lstlisting}
}
\end{AIbox} 
\caption{ \textbf{Few-shot Examples for the ARC-Challenge.} For a given shot $x$, the $x\!-\!shot$ utilizes the first $x$ examples.}
\label{fig:arc_7shot}
\vspace{-5mm}
\end{figure*}

\begin{figure*}[!ht] 
\vspace{-5mm}
\begin{AIbox}{Few-shot examples for the CommonsenseQA}
{
{\color{blue}\bf Example 1:}
\begin{lstlisting}[style=code]
The sanctions against the school were a punishing blow, and they seemed to what the efforts the 
school had made to change?
A. ignore
B. enforce
C. authoritarian
D. yell at
E. avoid
Answer: A
\end{lstlisting}
{\color{blue}\bf Example 2:}
\begin{lstlisting}[style=code]
Sammy wanted to go to where the people were. Where might he go?
A. race track
B. populated areas
C. the desert
D. apartment
E. roadblock
Answer: B
\end{lstlisting}
{\color{blue}\bf Example 3:}
\begin{lstlisting}[style=code]
To locate a choker not located in a jewelry box or boutique where would you go?
A. jewelry store
B. neck
C. jewlery box
D. jewelry box
E. boutique
Answer: A
\end{lstlisting}
{\color{blue}\bf Example 4:}
\begin{lstlisting}[style=code]
Google Maps and other highway and street GPS services have replaced what?
A. united states
B. mexico
C. countryside
D. atlas
E. oceans
Answer: D
\end{lstlisting}
{\color{blue}\bf Example 5:}
\begin{lstlisting}[style=code]
The fox walked from the city into the forest, what was it looking for?
A. pretty flowers.
B. hen house
C. natural habitat
D. storybook
E. dense forest
Answer: C
\end{lstlisting}
{\color{blue}\bf Example 6:}
\begin{lstlisting}[style=code]
What home entertainment equipment requires cable?
A. radio shack
B. substation
C. cabinet
D. television
E. desk
Answer: D
\end{lstlisting}
{\color{blue}\bf Example 7:}
\begin{lstlisting}[style=code]
The only baggage the woman checked was a drawstring bag, where was she heading with it?
A. garbage can
B. military
C. jewelry store
D. safe
E. airport
Answer: E
\end{lstlisting}
}
\end{AIbox} 
\caption{ \textbf{Few-shot Examples for the CommonsenseQA.} For a given shot $x$, the $x\!-\!shot$ utilizes the first $x$ examples.}
\label{fig:csqa_7shot}
\vspace{-5mm}
\end{figure*}

\begin{figure*}[!ht] 
\begin{AIbox}{Prompt Template for Rewriting Prompts}
{
You are a helpful AI assistance for question rewriting, and you must adhere to the following rules while striving to create the best possible question.\\
\noindent\par
Please be aware that the following is the fundamental rules you *must* adhere to when rewriting contents:
\begin{lstlisting}[style=code]
```
- Make sure the semantics are exactly the same before and after the change.
- Do not alter the formatting requirements within the question.
- Do not modify any proper nouns, such as names of people or places.
- Do not return content that hasn't changed in any way.
- The content you generate must be authentic in expression and logical to the native speaker.
```
\end{lstlisting}
\noindent\par
The given question:
\begin{lstlisting}[style=code]
```
{question}
```
\end{lstlisting}
\noindent\par
Please directly generate the response in following JSON format:
\begin{lstlisting}[style=code]
```json
{{
    //The rewritten question. Note that you only need to rewrite the question, not answer it.
    Rewritten_question: String;
}}
```
\end{lstlisting}

}
\end{AIbox} 
\caption{\textbf{The Prompt Template for Guiding GPT-4 in Rewriting Prompts.}}
\label{fig:meta_prompt}
\vspace{-5mm}
\end{figure*}
% This is an appendix.

\end{document}